\documentclass[11pt]{article}

\usepackage{acl}

\usepackage{times}
\usepackage{latexsym}
\usepackage[T1]{fontenc}
\usepackage[utf8]{inputenc}
\usepackage{microtype}
\usepackage{graphicx}
\usepackage{amsmath,amssymb}
\usepackage{float}
\usepackage{booktabs}
\usepackage{algorithm}
\usepackage{algorithmic}
\usepackage{tabularx}
\usepackage{array}
\usepackage{tablefootnote}
\usepackage{subcaption}
\usepackage{xcolor}
\usepackage{pifont}
\usepackage{makecell}
\usepackage{multirow}
\usepackage{adjustbox}
\usepackage{placeins}

\newcommand{\cmark}{\textcolor{green!60!black}{\ding{51}}}
\newcommand{\xmark}{\textcolor{red!90!black}{\ding{55}}}

\newcommand{\cc}[1]{\textcolor{black}{#1}}

\newcommand{\score}[2]{\ensuremath{#1\,{\scriptstyle \pm #2}}}
\newcommand{\scorebf}[2]{\ensuremath{\mathbf{#1}\,{\scriptstyle \pm #2}}}

\newcommand{\objcell}[1]{%
\begin{minipage}[t]{\linewidth}
\vspace{-2.0em}
\begin{equation*}
#1
\end{equation*}
\vspace{-1.4em}
\end{minipage}
}

\newcommand{\eg}{\emph{e.g.,}~}

\usepackage{pifont}

\usepackage{amsmath} 
\usepackage{bm} 
\usepackage{bbm} 
\usepackage{amsthm}
\usepackage{mathtools}

\newcommand{\ep}{\varepsilon}
\renewcommand{\epsilon}{\ep}

\makeatletter
\@ifundefined{proposition}{%
    
}{}
\@ifundefined{definition}{%
}{}
\@ifundefined{lemma}{%
    
}{}
\@ifundefined{corollary}{%
    \newtheorem{corollary}{Corollary}
}{}
\@ifundefined{theorem}{%
    
}{}
\@ifundefined{corollary}{%
    
}{}
\@ifundefined{assumption}{%
    
}{}
\@ifundefined{problem}{%
    
}{}
\@ifundefined{problem*}{%
    \newtheorem*{problem*}{Problem}
}{}
\@ifundefined{claim}{%
    
}{}
\@ifundefined{example}{%
    
}{}
\@ifundefined{implication}{%
    
}{}
\@ifundefined{remark*}{%
    \newtheorem*{remark*}{Remark}
}{}
\makeatother

\newtheorem{innercustomgeneric}{\customgenericname}
\providecommand{\customgenericname}{}
\newcommand{\newcustomtheorem}[2]{%
  \newenvironment{#1}[1]
  {%
   \renewcommand\customgenericname{#2}%
   \renewcommand\theinnercustomgeneric{##1}%
   \innercustomgeneric
  }
  {\endinnercustomgeneric}
}
\newcustomtheorem{customclaim}{Claim}

\renewcommand{\[}						{\left[}

\def\Ds{\mathcal{{D}}}

\def\Ls{\mathcal{{L}}}

\def\Xs{\mathcal{{X}}}
\def\Ys{\mathcal{{Y}}}

\title{SAUL: Sharpness-Aware Augmented-Lagrangian Unlearning}

\author{
  \textbf{Jaewan Choi\textsuperscript{1*}},
  \textbf{Junyoung Yang\textsuperscript{1*}},
  \textbf{Sangdon Park\textsuperscript{1,2}} \\
  \textsuperscript{1}Computer Science and Engineering, POSTECH \\
  \textsuperscript{2}Graduate School of Artificial Intelligence, POSTECH \\
  \texttt{\{cjw7656, sheepjun0330, sangdon\}@postech.ac.kr} \\
  \textsuperscript{*}Equal contribution.
}

\begin{document}

\maketitle

\begin{abstract}
Machine unlearning in Large Language Models (LLMs) faces a critical trade-off between erasing target knowledge and preserving general utility.
We propose \textbf{SAUL} (Sharpness-Aware Augmented-Lagrangian Unlearning), which formulates unlearning as a constrained minimization problem following the principle of ``\emph{forget enough, but no more than necessary.}'' 
At its core, SAUL formulates forgetting as an explicit constraint with a prescribed satisfaction criterion, whereas prior unlearning methods typically specify the desired level of forgetting implicitly through optimization objectives.
An augmented Lagrangian controller adaptively adjusts forget-side pressure according to constraint violation and can eventually deactivate the forget-side update as the prescribed criterion remains satisfied.
Sharpness-aware updates on both retain and forget objectives, together with a dual-optimizer design that maintains role-separated states, further stabilize the resulting unlearning dynamics.
We evaluate SAUL on the TOFU, WMDP, and MUSE benchmarks, demonstrating favorable forgetting--utility trade-offs over representative sharpness- and perturbation-based baselines under benchmark-specific forgetting criteria.
\cc{
Beyond the complete SAUL framework, we further show on TOFU that applying the augmented-Lagrangian controller as a \emph{drop-in modifier} to representative baselines improves their post-forgetting utility, demonstrating the practical value of explicit forgetting control.
}
\end{abstract}

\section{Introduction}

Large language models (LLMs) are trained on massive corpora that often contain sensitive, copyrighted, or harmful information~\cite{carlini2021extracting,eldan2023harry_potter,li2024wmdp}.
\textbf{LLM unlearning} has emerged as a critical framework for addressing these concerns by enabling models to ``forget'' specific data without the prohibitive cost of retraining from scratch~\cite{yao2024largelanguagemodelunlearning,liu2025rethinking}.
A standard unlearning task must erase target knowledge from a \emph{forget set} while preserving general capabilities on a \emph{retain set}.
Many existing methods handle this conflict through weighted objectives that combine forget and retain terms~\cite{zhang2024npo,fan2025relearning,bu2025unlearningmultitaskoptimizationnormalized}, but their trade-off coefficients are sensitive to dataset, model architecture, and forgetting ratio~\cite{maini2024tofu,zhang2024npo}.

\begin{figure}[t]
    \centering
    \vspace{-1.0em}
    \includegraphics[width=0.85\linewidth]{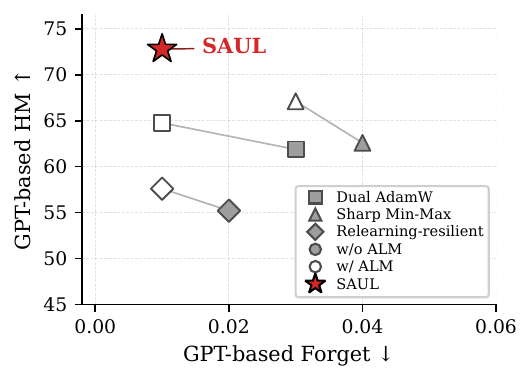}
    \caption{GPT-based utility--forgetting trade-off on ToFU with $1\%$ forget ratio.
    Better methods lie toward the upper-left, with higher HM and lower Forget.
    Gray lines connect each baseline to its ALM-enhanced variant.
    SAUL achieves the highest HM while maintaining a low Forget score.}
    \label{fig:hm_forget_tradeoff}
    \vspace{-0.5em}
\end{figure}

\begin{figure}[t]
    \centering
    \vspace{-1.0em}
    \includegraphics[width=0.85\linewidth]{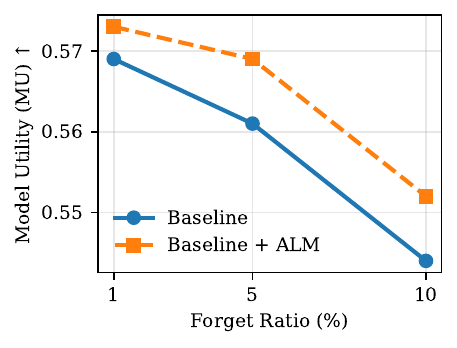}
    \vspace{-1.0em}
    \caption{Model utility under an ALM constraint (Dual AdamW baseline). Across all forget ratios, ALM improves model utility over the standard Dual AdamW optimization.}
    \label{fig:motivation_experiment}
    \vspace{-0.8em}
\end{figure}

Recent optimization-based approaches, including bi-level unlearning methods, introduce more structured ways to handle the forget--retain conflict by separating the roles of forgetting and retention across nested or sequential objectives~\cite{reisizadeh2025blurbileveloptimizationapproach,asif2025ofmuoptimizationdrivenframeworkmachine}.
A parallel line of work mitigates the instability of aggressive forgetting through sharpness- or perturbation-aware optimization~\cite{fan2025relearning,tang2025sharpness_aware_mu,kim2025uam,malekmohammadi2025sharpness_param_selection}, building on Sharpness-Aware Minimization (SAM)~\cite{foret2021sam} to obtain solutions stable under local parameter perturbations. While these approaches improve expressiveness and robustness, they still treat forgetting as open-ended \emph{optimization pressure} rather than an explicit constraint with a prescribed satisfaction criterion, allowing unnecessary forgetting to erode retain-side utility.

In this paper, we propose \textbf{Sharpness-Aware Augmented Lagrangian Unlearning (SAUL)}, a framework that casts LLM unlearning as a constrained minimization problem following the principle of \emph{forgetting enough, but no more than necessary}.
Unlike weighted formulations, SAUL uses a threshold $\alpha$ to specify a target forgetting level, allowing optimization to focus on retain-side utility once the criterion is satisfied.

At the core of SAUL is an \emph{augmented Lagrangian} controller~\cite{Hestenes1969MultiplierAG,powell1969method} that adaptively activates or releases forgetting pressure depending on whether the criterion is met.
As the prescribed criterion remains satisfied, the projected multiplier decreases and can eventually reach zero and the forget-side gradient is gated off entirely, so optimization focuses solely on preserving retain-side utility---an \emph{adaptive deactivation} mechanism absent from prior constrained or penalty-based unlearning methods.
To stabilize the resulting dynamics, SAUL further combines sharpness-aware updates on both retain and forget objectives with a \emph{dual} optimizer design~\citep{zhong2025dualoptimenhancingefficacystability} that maintains role-separated states for retention and forgetting, reducing interference between their adaptive moments.

We evaluate SAUL across {\sc ToFU}, WMDP, and MUSE under matched forgetting criteria, where SAUL improves the forgetting--utility trade-off over representative unlearning baselines. Beyond SAUL itself, Figures~\ref{fig:hm_forget_tradeoff} and~\ref{fig:motivation_experiment} show that the augmented Lagrangian component, applied as a drop-in modifier to existing baselines, consistently improves post-forgetting utility, highlighting the broader value of explicit forgetting control.

Our contributions are summarized as follows:
\begin{itemize}
    \item We formulate LLM unlearning as a constrained minimization problem in which a target threshold $\alpha$ explicitly specifies a prescribed satisfaction level for forgetting, and propose an augmented-Lagrangian controller that reduces and can eventually deactivate forget-side pressure as the criterion remains satisfied.
    
    \item We show that the proposed controller can be integrated into representative sharpness- and perturbation-based unlearning methods. Applied as a drop-in modifier on {\sc ToFU}, it consistently improves their post-forgetting utility, demonstrating the practical benefit of explicit forgetting control.
    
    \item We instantiate the framework as \textbf{SAUL}, which integrates the controller with sharpness-aware updates and role-separated dual optimizer states, and show that it achieves favorable forgetting--utility trade-offs across {\sc ToFU}, WMDP, and MUSE under benchmark-specific forgetting criteria.
\end{itemize}

\section{Related Work}\label{sec:rel}

\paragraph{LLM unlearning and weighted objectives.}
Many LLM unlearning methods formulate forgetting and retention as a scalarized problem, combining forget- and retain-side objectives through a trade-off coefficient. Representative examples include gradient ascent, gradient-difference objectives, loss-adjustment variants, and preference-optimization formulations~\cite{jang2023knowledge_unlearning,wang2024llmunlearninglossadjustment,bu2025unlearningmultitaskoptimizationnormalized,zhang2024npo,mekala2024alternatepreferenceoptimizationunlearning}. While effective, these methods specify the desired forgetting level implicitly through coefficients that depend on the dataset, model, and forget ratio~\cite{maini2024tofu,zhang2024npo}, potentially causing insufficient forgetting or unnecessary utility degradation.

\paragraph{Bi-level and structured optimization for unlearning.}
\cc{Recent work has explored structured formulations beyond scalarization. BLUR~\cite{reisizadeh2025blurbileveloptimizationapproach} casts unlearning as a bi-level problem whose lower-level objective prioritizes forgetting and upper-level objective preserves retain-side utility. OFMU~\cite{asif2025ofmuoptimizationdrivenframeworkmachine} uses an inner forgetting maximization with a similarity-aware penalty and an outer utility-restoration minimization. These methods separate the roles of forgetting and retention, but still lack a prescribed threshold that determines when forgetting is sufficient.}

\paragraph{Constrained forgetting and augmented Lagrangian methods.}
Augmented Lagrangian methods combine dual-variable updates with quadratic penalties to enforce constraints without relying solely on a fixed penalty coefficient~\cite{Hestenes1969MultiplierAG,bertsekas1982constrained,nocedal2006numerical}. Existing constrained LLM unlearning methods place the constraint on the \emph{retain side}: Constrained Entropic Unlearning~\cite{entesari2025constrainedentropicunlearningprimaldual} uses a hard retain constraint, while Cheng et al.~\cite{entanglement_al_openreview} constrain the retain-side performance of $\theta$ relative to the original model $\theta_0$. In both formulations, the forget-side objective remains active. SAUL instead constrains the \emph{forget side}; as the criterion remains satisfied, the projected multiplier decreases and can eventually reach zero, gating off the forget-side update. We further evaluate the controller as a drop-in modifier to representative sharpness- and perturbation-based objectives on {\sc ToFU} (Section~\ref{sec:alm-augmented}).

\paragraph{Sharpness-aware and role-separated optimization.}
Sharpness-Aware Minimization (SAM) promotes solutions stable under local weight perturbations~\cite{foret2021sam,bahri2022sharpnessawareminimizationimproveslanguage}. Sharpness- and perturbation-aware unlearning methods use this principle to stabilize retention or forgetting behavior and reduce susceptibility to relearning-style recovery~\cite{fan2025relearning,tang2025sharpness_aware_mu,kim2025uam,malekmohammadi2025sharpness_param_selection}, but generally lack an explicit criterion for forgetting sufficiency. Separately, dual-optimizer methods reduce interference by maintaining distinct optimizer states for competing objectives~\cite{zhong2025dualoptimenhancingefficacystability}. SAUL uses these ideas as \emph{supporting components}, while its central contribution is explicit forget-side constraint control.
\section{Problem: Unlearning with Constraints}
\label{sec:problem}

Machine unlearning aims to remove the influence of a designated subset of training data (a \emph{forget set}) from a model, while preserving (and potentially improving) performance on the remaining data (a \emph{retain set}).
We view unlearning as a constrained minimization problem.
In particular, we consider model parameters $\theta \in \mathbb{R}^d$, an example space $\Xs$, a label space $\Ys$, a \emph{retain (multi-)set} $\Ds_r \subseteq \Xs \times \Ys$, a \emph{forget (multi-)set} $\Ds_f \subseteq \Xs \times \Ys$, and a per-example training loss $\ell(\theta,x,y)$ (\eg cross-entropy).
To measure how well the model fits each set, we define the \emph{retain loss}
$\mathcal{L}_r(\theta) \coloneqq \frac{1}{|\Ds_r|} \sum_{(x,y)\in \Ds_r} \ell(\theta,x,y)$
as the average per-example loss on the retain set, and similarly the \emph{forget loss}
$\mathcal{L}_f(\theta) \coloneqq \frac{1}{|\Ds_f|} \sum_{(x,y)\in \Ds_f} \ell(\theta,x,y)$
on the forget set.
Our goal is to address the following constrained minimization:
\begin{equation}
    \label{eq:goal}
    \min_\theta \Ls_r(\theta) \quad \text{subj. to} \quad \Ls_f(\theta) \ge \alpha,
\end{equation}
where $\alpha$ is a user-specified threshold that defines an explicit satisfaction level for the chosen forget-side training measure, rather than specifying the forgetting--retention balance implicitly through a trade-off coefficient.
The constraint defines sufficient forgetting in terms of the chosen forget-side loss by requiring $\mathcal{L}_f(\theta)\ge\alpha$, while the retain objective discourages utility degradation beyond what is necessary to satisfy this criterion.
Moreover, minimizing the retain loss remains compatible with cases where removing the forget set also improves performance on the remaining data. In Appendix~\ref{sec:threshold-calibration}, we introduce a margin-based certified instantiation
in which the threshold $\alpha$ admits a direct prediction-level
interpretation, and explain how the resulting certificate can be
used post hoc to evaluate models trained with the
cross-entropy-based constraint.


    



\section{SAUL: Sharpness-Aware Augmented Lagrangian Unlearning}
\label{sec:method}

\begin{figure*}[t]
    \centering
    \includegraphics[width=\textwidth]{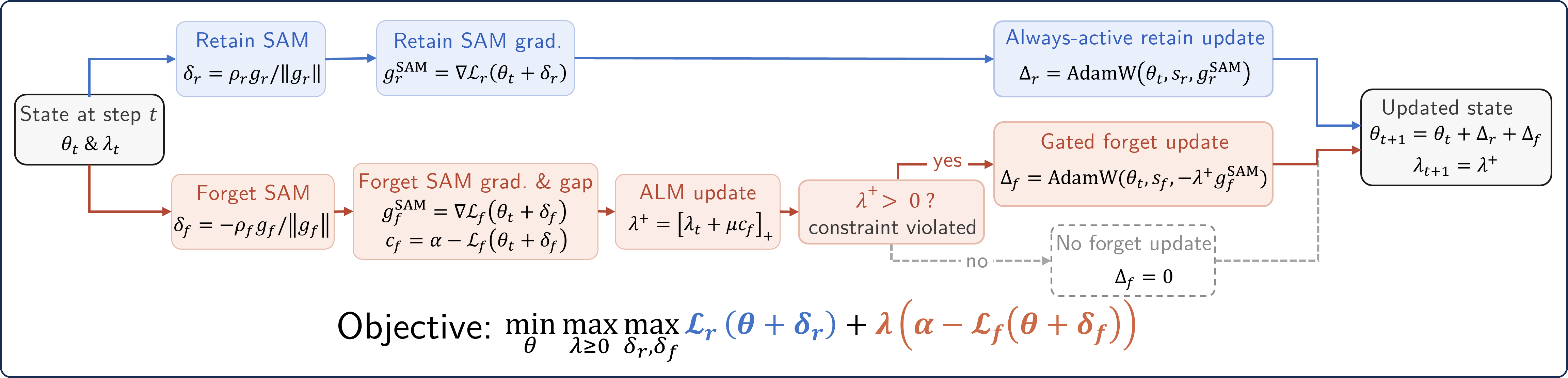}
    \caption{Overview of the SAUL update. The retain branch is always active and applies a SAM-based update to preserve retain-side utility. The forget branch computes a SAM-based forget gradient and an augmented-Lagrangian constraint gap, updates $\lambda^+=[\lambda_t+\mu c_f]_+$, and applies the forget update only when $\lambda^+>0$. Thus, forgetting is activated only when the constraint is violated, while retain-side optimization remains active throughout training.}
    \label{fig:example_wide}
\end{figure*}

Building on the constrained formulation in Section \ref{sec:problem}, we propose
\emph{Sharpness-Aware Augmented Lagrangian Unlearning} (\textbf{SAUL}),
which combines an augmented-Lagrangian constraint with sharpness-aware updates and dual optimizer states
to decouple the dynamics of forgetting and retention.

\subsection{Augmented Lagrangian Method}
\label{sec:alm}

Here, we apply the augmented Lagrangian method (ALM) \cite{Hestenes1969MultiplierAG, powell1969method} to unlearning. ALM converts the constrained primal problem~(\ref{eq:goal}) into the following unconstrained primal problem with a Lagrangian multiplier:
\begin{equation}
    \label{eq:primalwithlag}
    \min_{\theta} \max_{\lambda \ge 0}  \Ls_r(\theta) + \lambda (\alpha - \Ls_f(\theta)),
\end{equation}
which is equivalent to~(\ref{eq:goal}). However, repeated constraint violations can make the multiplier grow and induce unstable primal updates.
ALM can be written as a proximal maximization over the next multiplier estimate $\lambda^+$, where the quadratic term prevents the dual variable from moving too far from the current estimate $\lambda$.
This gives the following regularized saddle problem:
\begin{equation}
    \label{eq:am:iter1}
    \min_{\theta} \max_{\lambda^+ \ge 0} \big[ \Ls_r(\theta) + \lambda^+ (\alpha - \Ls_f(\theta)) - \frac{1}{2 \mu} \| \lambda^+  - \lambda \|^2 \big],
\end{equation}
where $\mu > 0$ is the ALM penalty parameter. Since the objective is concave in $\lambda^+$, it admits the closed-form solution:
\begin{equation*}
    \lambda^+ = \big[ \lambda + \mu ( \alpha - \Ls_f(\theta) ) \big]_+.
\end{equation*}
Let
$
c_f := \alpha - \Ls_f(\theta)
$
denote the forget-side constraint violation, so that the constraint $\Ls_f(\theta) \ge \alpha$ is equivalent to $c_f \le 0$.
Plugging the closed-form multiplier update back into~(\ref{eq:am:iter1}), the primal update takes the following two-case form:
\begin{equation*}
\min_{\theta}
\begin{cases}
\Ls_r(\theta)
+ \lambda c_f
+ \frac{\mu}{2} c_f^2,
& \lambda^+ > 0, \\
\Ls_r(\theta),
& \lambda^+ = 0.
\end{cases}
\end{equation*}
When $\lambda^+ > 0$, the augmented term penalizes violations of the forget-side constraint.
When the projection yields $\lambda^+ = 0$, the update reduces to retain-loss minimization.
For both cases, the updated $\theta$ may not satisfy the constraint, so we iterate this procedure by returning to~(\ref{eq:am:iter1}) with $\lambda \leftarrow \lambda^+$.

Crucially, as constraint violations disappear, the projected multiplier can decrease and eventually deactivate the forget-side term.
When $\lambda^+ = 0$, optimization focuses solely on minimizing the retain loss, avoiding unnecessary forgetting pressure.

\subsection{Sharpness-Aware Augmented Lagrangian Method}
\label{sec:sam}

Sharpness-Aware Minimization (SAM) \cite{foret2021sam} biases optimization toward parameter regions where the loss landscape is flat under local weight perturbations, by approximately solving an inner min--max problem. We extend this idea to both retention and forgetting, so that the resulting solutions remain stable under local weight perturbations on each side.

To this end, we consider a sharpness-aware retain loss
\begin{equation*}
    \Ls_r^{\rho_r}(\theta) \coloneqq \max_{\| \delta_r \| \le \rho_r} \Ls_r(\theta + \delta_r)
\end{equation*}
instead of $\Ls_r(\theta)$. Minimizing this worst-case retain loss yields parameters where the retain landscape is flat, so retain performance remains stable under small weight changes. For the forget side, we consider a sharpness-aware forget loss in the opposite direction:
\begin{equation*}
    \Ls_f^{\rho_f}(\theta) \coloneqq \min_{\| \delta_f \| \le \rho_f} \Ls_f(\theta + \delta_f)
\end{equation*}
instead of $\Ls_f(\theta)$. Here, the inner minimization finds the perturbation \emph{most favorable to recovering forget knowledge}, and maximizing this best-case forget loss forces the forget landscape to remain high even at the most easily recoverable point in the neighborhood. Note the asymmetry: the retain side uses inner max so that the worst case stays low, while the forget side uses inner min so that the best case stays high. The full sharpness-aware augmented Lagrangian method follows ALM in Section~\ref{sec:alm}, but replaces $\Ls_r(\theta)$ and $\Ls_f(\theta)$ with $\Ls_r^{\rho_r}(\theta)$ and $\Ls_f^{\rho_f}(\theta)$, respectively. See Algorithm~\ref{alg:main} for our complete algorithm.

Combining the sharpness-aware losses with the ALM formulation in
(\ref{eq:primalwithlag}) yields the following robust min--max problem:
\begin{equation}
    \label{eq:sharp_alm_minmax}
    \min_{\theta} \max_{\lambda \ge 0} \max_{\delta_r, \delta_f}
    \Ls_r(\theta + \delta_r)
    + \lambda \big(\alpha - \Ls_f(\theta + \delta_f)\big).
\end{equation}
For the forget term, since $\lambda \ge 0$, maximizing
$-\lambda \Ls_f(\theta+\delta_f)$ over $\delta_f$ is equivalent to minimizing
$\Ls_f(\theta+\delta_f)$, which matches the sharpness-aware forget loss.
Because $\delta_r=0$ and $\delta_f=0$ are feasible perturbations,
(\ref{eq:sharp_alm_minmax}) upper-bounds the ALM objective in
(\ref{eq:primalwithlag}) for any fixed $\theta$ and $\lambda$.
In practice, as in standard SAM, we approximate each inner problem with a single first-order perturbation step. Minimizing this upper bound therefore controls the original ALM objective while encouraging stability under local weight perturbations on both sides.
This gives a principled interpretation of the forget-side sharpness step as approximating a robust forgetting constraint that should remain satisfied even under local weight perturbations. We emphasize that this robustness is restricted to weight-space perturbations and does not by itself imply resistance to prompt-level adversarial attacks or recovery attacks.

\subsection{Dual Optimizer States}
\label{sec:dual}

While the augmented Lagrangian formulation in Section~\ref{sec:alm} specifies when the forgetting constraint should be enforced through the multiplier $\lambda$, and the sharpness-aware losses in Section~\ref{sec:sam} improve the \emph{robustness} of the retain/forget objectives, a remaining source of instability lies in the \emph{optimizer dynamics} used to carry out these updates \citep{zhong2025dualoptimenhancingefficacystability}.

A common implementation optimizes the (sharpness-aware) ALM objective with a \emph{single} adaptive optimizer (e.g., AdamW), which couples forgetting and retention through shared first- and second-moment estimates. In unlearning, however, retain-side and forget-side gradients often have conflicting semantics and markedly different scales---utility-preserving descent versus constraint-enforcing updates---and mixing such heterogeneous signals in one state can lead to optimizer-state interference, where accumulated moments become inconsistent with either role.

To mitigate this, SAUL maintains \emph{two independent optimizer states} for the \emph{same} model parameters~$\theta$: a forgetting optimizer state $s_f$ and a retention optimizer state $s_r$ (both based on AdamW in our implementation). The parameters remain shared; only the optimizer statistics are decoupled. $s_f$ is updated using forget-side gradients only, while $s_r$ is updated using retain-side gradients only, so each state accumulates objective-consistent moments suitable for its role. At each iteration, SAUL first computes both $g_r^{\mathrm{SAM}}$ and $g_f^{\mathrm{SAM}}$ at the current parameters $\theta_t$.
It then applies the forget-side update only when $\lambda^+ > 0$, followed by the retain-side update using role-separated optimizer states. Algorithm~\ref{alg:main} summarizes the complete procedure, combining (i) ALM-based adaptive activation of forgetting, (ii) sharpness-aware gradients for robust optimization, and (iii) role-separated optimizer states for stable unlearning dynamics.


\subsection{ALM-Augmented Variants}
\label{sec:alm-augmented}

Beyond SAUL itself, the augmented-Lagrangian controller can also be applied as a drop-in modifier to existing unlearning objectives.
Given a baseline objective $\mathcal{J}(\theta)$ and a forgetting measure $F(\theta)$, where larger $F(\theta)$ indicates stronger forgetting, we impose the constrained form
\begin{equation*}
\min_{\theta}\ \mathcal{J}(\theta)
\quad \text{s.t.}\quad
F(\theta)\ge \alpha .
\end{equation*}
ALM introduces a dual variable $\lambda\ge0$ and a penalty parameter $\mu>0$, with the projected update
\begin{equation*}
\lambda^{+}\leftarrow \big[\lambda + \mu(\alpha - F(\theta))\big]_+ .
\end{equation*}
This update increases forget-side pressure when the constraint is violated and reduces it as the constraint remains satisfied, replacing fixed trade-off tuning with constraint-driven adaptation.

Unlike SAUL, which integrates ALM with sharpness-aware updates on both retain and forget sides plus role-separated optimizer states, these ALM-augmented baselines apply only the constraint mechanism on top of the original baseline structure.
These ALM-augmented variants allow us to isolate the effect of constraint-driven forgetting control from the underlying sharpness-aware objectives.
We evaluate the post-forgetting utility of these ALM-augmented variants in Section~\ref{sec:exp}, with detailed baseline-specific objectives provided in Appendix~\ref{app:alm_variants}.

\section{Experiments}\label{sec:exp}

We evaluate SAUL on three LLM unlearning benchmarks—ToFU~\cite{maini2024tofu}, WMDP~\cite{li2024wmdp}, and MUSE~\cite{shi2024muse}—covering author-specific fictitious forgetting, hazardous knowledge removal, and verbatim memorization.
All methods are tuned under matched forgetting criteria so that comparisons reflect the forgetting--utility trade-off at comparable levels of forgetting.
Detailed training configurations and hyperparameters are provided in Appendix~\ref{app:experimental_setup}.

\subsection{Setup}
\label{sec:exp_setup}

\paragraph{Benchmarks.}
We evaluate methods across three benchmark-specific settings, following the established model and evaluation configurations used for each benchmark rather than adopting a single shared backbone. For \textbf{ToFU}, we use \texttt{Llama-3.2-1B-Instruct} as the main backbone under the standard question--answering setting with forgetting ratios of $1\%$, $5\%$, and $10\%$. We also evaluate GPT-paraphrased questions to test robustness to query rephrasing; their construction is described in Appendix~\ref{app:paraphrased_tofu}, and the corresponding results are reported in Appendix~\ref{app:additional_results}. Additional \texttt{Llama-3.2-3B-Instruct} results are reported in Appendix~\ref{app:additional_results}. For \textbf{WMDP}, we follow its benchmark-specific setup using \texttt{Zephyr-7B-$\beta$} and unlearn WMDP-Bio and WMDP-Cyber, evaluating hazardous-knowledge accuracy and MMLU utility. For \textbf{MUSE}, we follow the benchmark-specific \texttt{Llama-2-7B} setup and use the standard VerbMem, KnowMem, and PrivLeak evaluation protocol. We report MUSE Books as the main evaluation setting and provide additional results on MUSE News in Appendix~\ref{app:muse_additional}.





\paragraph{Baselines.}
We compare against representative sharpness- and perturbation-based unlearning methods: the relearning-resilient sharpness-aware baseline of Fan et al.~\citep{fan2025relearning} and Sharp Min--Max~\citep{tang2025sharpness_aware_mu} without masking.
For brevity, we refer to the former as Relearning-resilient in tables and result discussions.
We also include Single AdamW, Dual AdamW~\citep{zhong2025dualoptimenhancingefficacystability}, and BLUR~\citep{reisizadeh2025blurbileveloptimizationapproach}. 
On ToFU, we additionally compare against Primal-Dual Unlearning (PDU)~\citep{entesari2025constrainedentropicunlearningprimaldual}, a recent retain-side constrained method that optimizes forgetting subject to an explicit retain-utility constraint.
For baselines that admit a comparable operating point, we further evaluate a ``+\,ALM'' variant to isolate the effect of explicit forgetting control.
Detailed baseline definitions and implementation details are provided in Appendix~\ref{app:baselines}.


\newcommand{\meanpm}[2]{#1 {\scriptsize $\pm$ #2}}

\begin{table*}[tb]
\centering
\small
\setlength{\tabcolsep}{3.0pt}
\renewcommand{\arraystretch}{0.95}
\caption{Unlearning performance with forget ratio =  $1\%$.
All methods are tuned to achieve Forget ROUGE close to 0.03 when possible.
\textbf{Bold} and \underline{underlined} values denote the best and second-best results for each metric, respectively.
\textcolor{red}{Red} values indicate cases where unlearning is not sufficiently achieved; methods with red Forget values are excluded from the best/second-best ranking.
A dash (--) indicates that the metric is not applicable under our hyperparameter tuning protocol.}
\begin{tabular}{@{}l || cc|ccc|cc@{}}
\toprule
Method
& \multicolumn{2}{c|}{\textbf{Automatic}}
& \multicolumn{5}{c}{\textbf{GPT-based}} \\
\cmidrule(r){2-3}\cmidrule(l){4-8}
& MU$\uparrow$ & $F_{\text{ROUGE}}\downarrow$
& Retain$\uparrow$
& World Facts$\uparrow$
& Real Authors$\uparrow$
& HM $\uparrow$
& Forget$\downarrow$ \\
\midrule
    
Original Model
& 0.60 & 0.87 & 90.0 & 80.3 & 81.4 & 81.9 & 82.5 \\

\midrule

Single AdamW
& \meanpm{0.45}{0.00} & \underline{\meanpm{0.02}{0.03}}
& \meanpm{50.45}{0.72}
& \meanpm{70.43}{0.97}
& \meanpm{35.20}{2.39}
& \meanpm{48.01}{1.50}
& \meanpm{0.17}{0.08} \\

Dual AdamW
& \meanpm{0.57}{0.00} & \underline{\meanpm{0.02}{0.00}}
& \meanpm{64.25}{0.56}
& \meanpm{80.00}{1.43}
& \meanpm{49.00}{1.58}
& \meanpm{61.87}{0.87}
& \meanpm{0.03}{0.02} \\

Dual AdamW + ALM
& \meanpm{0.57}{0.00} & \textbf{\meanpm{0.01}{0.00}}
& \meanpm{63.20}{0.27}
& \textbf{\meanpm{82.56}{1.43}}
& \meanpm{54.40}{1.14}
& \meanpm{64.76}{0.68}
& \textbf{\meanpm{0.01}{0.01}} \\

Sharp Min--Max
& \meanpm{0.56}{0.00} & \underline{\meanpm{0.02}{0.00}}
& \meanpm{53.45}{0.37}
& \meanpm{81.54}{0.97}
& \meanpm{59.00}{1.58}
& \meanpm{62.59}{0.58}
& \meanpm{0.04}{0.02} \\

Sharp Min--Max + ALM
& \underline{\meanpm{0.58}{0.01}} & \textbf{\meanpm{0.01}{0.00}}
& \meanpm{58.05}{0.37}
& \meanpm{80.85}{0.76}
& \underline{\meanpm{66.20}{0.84}}
& \underline{\meanpm{67.11}{0.25}}
& \meanpm{0.03}{0.03} \\

Relearning-resilient
& \meanpm{0.51}{0.01} & \textbf{\meanpm{0.01}{0.00}}
& \meanpm{46.40}{0.38}
& \meanpm{77.78}{0.60}
& \meanpm{50.20}{0.84}
& \meanpm{55.21}{0.23}
& \underline{\meanpm{0.02}{0.02}} \\

Relearning-resilient + ALM 
& \meanpm{0.52}{0.01} & \underline{\meanpm{0.02}{0.01}}
& \meanpm{49.90}{0.29}
& \meanpm{80.00}{1.43}
& \meanpm{51.20}{0.84}
& \meanpm{57.60}{0.15}
& \textbf{\meanpm{0.01}{0.01}} \\

BLUR
& \meanpm{0.57}{0.14} & \meanpm{0.28}{0.20}
& \meanpm{63.00}{1.40}
& \meanpm{80.00}{3.40}
& \meanpm{73.60}{4.00}
& \meanpm{71.40}{2.50}
& \textcolor{red}{\meanpm{36.00}{2.80}} \\

PDU
& \meanpm{0.56}{0.02} & \meanpm{0.17}{0.02}
& \meanpm{60.42}{1.04}
& \meanpm{79.2}{0.99}
& \meanpm{62.00}{2.00}
& \meanpm{66.20}{0.70}
& \textcolor{red}{\meanpm{8.33}{3.82}} \\

\midrule
\textbf{SAUL}
& \textbf{\meanpm{0.59}{0.01}} & \underline{\meanpm{0.02}{0.02}}
& \textbf{\meanpm{68.25}{1.09}}
& \meanpm{79.66}{1.27}
& \textbf{\meanpm{71.40}{0.55}}
& \textbf{\meanpm{72.79}{0.54}}
& \textbf{\meanpm{0.01}{0.01}} \\

w/o ALM
& \meanpm{0.57}{0.01} & \textbf{\meanpm{0.01}{0.01}}
& \underline{\meanpm{64.95}{0.60}}
& \meanpm{78.29}{1.55}
& \meanpm{41.40}{1.14}
& \meanpm{57.32}{0.48}
& \underline{\meanpm{0.02}{0.02}} \\

w/o SAM
& \meanpm{0.55}{0.01} & \textbf{\meanpm{0.01}{0.01}}
& \meanpm{64.50}{0.68}
& \underline{\meanpm{82.39}{1.30}}
& \meanpm{56.20}{1.92}
& \meanpm{66.01}{1.01}
& \underline{\meanpm{0.02}{0.01}} \\

w/o ALM + SAM
& \meanpm{0.57}{0.00} & \textbf{\meanpm{0.01}{0.01}}
& \meanpm{64.10}{1.11}
& \meanpm{80.51}{1.64}
& \meanpm{47.60}{1.14}
& \meanpm{61.17}{0.39}
& \meanpm{0.16}{0.33} \\

\bottomrule
\end{tabular}
\label{tab:forget1}
\end{table*}

\begin{table*}[tb]
\centering
\small
\setlength{\tabcolsep}{3.0pt}
\renewcommand{\arraystretch}{0.95}
\caption{
Unlearning performance with forget ratio = $1\%$ under GPT-paraphrased questions.
All hyperparameters are selected on the original-question setting and then reused for paraphrased-question evaluation without additional tuning.
Notation follows Table~\ref{tab:forget1}.
}
\begin{tabular}{@{}l || cc|ccc|cc@{}}
\toprule
Method
& \multicolumn{2}{c|}{\textbf{Automatic}}
& \multicolumn{5}{c}{\textbf{GPT-based}} \\
\cmidrule(r){2-3}\cmidrule(l){4-8}
& MU$\uparrow$ & $F_{\text{ROUGE}}\downarrow$
& Retain$\uparrow$
& World Facts$\uparrow$
& Real Authors$\uparrow$
& HM $\uparrow$
& Forget$\downarrow$ \\
\midrule
    
Original Model
& 0.54 & 0.44 & 56.75 & 79.49 & 77.00 & 69.46 & 50.0\\

\midrule

Single AdamW
& \meanpm{0.49}{0.01} & \underline{\meanpm{0.03}{0.01}}
& \meanpm{46.46}{2.38}
& \meanpm{68.22}{9.94}
& \meanpm{25.60}{11.22}
& \meanpm{38.25}{8.79}
& \meanpm{2.00}{2.09} \\

Dual AdamW
& \underline{\meanpm{0.51}{0.02}} & \underline{\meanpm{0.03}{0.01}}
& \textbf{\meanpm{53.14}{2.09}}
& \meanpm{71.98}{7.90}
& \meanpm{47.20}{17.78}
& \meanpm{55.89}{16.37}
& \textbf{\meanpm{0.50}{1.12}} \\

Dual AdamW + ALM
& \textbf{\meanpm{0.53}{0.01}} & \meanpm{0.04}{0.00}
& \meanpm{52.62}{1.87}
& \underline{\meanpm{75.72}{5.01}}
& \meanpm{55.40}{10.60}
& \meanpm{59.20}{5.36}
& \meanpm{2.50}{0.00} \\

Sharp Min--Max
& \meanpm{0.47}{0.04} & \meanpm{0.12}{0.18}
& \meanpm{43.12}{3.51}
& \meanpm{70.10}{5.16}
& \meanpm{41.80}{5.93}
& \meanpm{48.74}{4.35}
& \meanpm{1.50}{2.24} \\

Sharp Min--Max + ALM
& \meanpm{0.50}{0.03} & \meanpm{0.05}{0.02}
& \meanpm{49.22}{1.74}
& \meanpm{73.86}{3.45}
& \meanpm{51.60}{8.50}
& \meanpm{56.14}{4.46}
& \meanpm{1.50}{1.37} \\

Relearning-resilient
& - & - & - & - & - & - & - \\

Relearning-resilient + ALM 
& - & - & - & - & - & - & - \\

BLUR
& \meanpm{0.53}{0.03} & \meanpm{0.44}{0.21}
& \meanpm{51.90}{3.30}
& \meanpm{78.80}{3.10}
& \meanpm{72.80}{4.50}
& \meanpm{65.60}{3.70}
& \textcolor{red}{\meanpm{49.00}{4.50}} \\

PDU
& \meanpm{0.50}{0.01} & \meanpm{0.34}{0.02}
& \meanpm{42.83}{2.36}
& \meanpm{74.93}{1.31}
& \meanpm{59.67}{1.15}
& \meanpm{56.09}{1.33}
& \textcolor{red}{\meanpm{36.67}{2.89}} \\

\midrule
\textbf{SAUL}
& \textbf{\meanpm{0.53}{0.01}} & \meanpm{0.13}{0.22}
& \meanpm{52.28}{1.40}
& \textbf{\meanpm{78.44}{2.29}}
& \textbf{\meanpm{63.60}{6.23}}
& \textbf{\meanpm{62.90}{2.18}}
& \meanpm{1.50}{1.37} \\

w/o ALM
& \meanpm{0.50}{0.03} & \underline{\meanpm{0.03}{0.02}}
& \meanpm{52.22}{2.53}
& \meanpm{70.94}{9.91}
& \meanpm{46.80}{24.77}
& \meanpm{51.44}{15.83}
& \underline{\meanpm{1.00}{2.24}} \\

w/o SAM
& \textbf{\meanpm{0.53}{0.01}} & \underline{\meanpm{0.03}{0.01}}
& \underline{\meanpm{53.12}{1.50}}
& \meanpm{75.23}{4.47}
& \underline{\meanpm{62.60}{12.99}}
& \underline{\meanpm{61.65}{4.73}}
& \underline{\meanpm{1.00}{1.37}} \\

w/o ALM + SAM
& \meanpm{0.50}{0.02} & \textbf{\meanpm{0.02}{0.01}}
& \meanpm{52.12}{1.62}
& \meanpm{68.02}{10.70}
& \meanpm{39.80}{16.30}
& \meanpm{49.05}{11.33}
& \textbf{\meanpm{0.50}{1.12}} \\

\bottomrule
\end{tabular}
\label{tab:forget1_qpara}
\end{table*}

\subsection{Results on ToFU}\label{sec:exp_tofu}

\paragraph{Evaluation metrics.}
We assess unlearning performance using two complementary automatic metrics from ToFU: \emph{model utility} (MU), which aggregates the probability assigned to the ground-truth answer, ROUGE-L recall, and the truth ratio via the harmonic mean; and \emph{Forget ROUGE} ($F_{\text{ROUGE}}$), which measures ROUGE-L recall between generated and ground-truth answers on the forget set. We additionally employ a GPT-based semantic evaluation that produces binary entailment judgments between generated and reference answers, applied to four subsets (\emph{Forget}, \emph{Retain}, \emph{Real Authors}, \emph{World Facts}); we report the harmonic mean (HM) of GPT-based scores on the latter three to summarize utility beyond forgetting.

\paragraph{Main results.}
Table~\ref{tab:forget1} reports results at forget ratio 1\%.
Since SAUL formulates forgetting as a constraint, we compare methods under the
matched-forgetting criterion $F_{\text{ROUGE}} \leq 0.03$.
SAUL satisfies this criterion with Forget ROUGE of 0.02 and achieves the best
harmonic mean (72.79), outperforming the next-best constraint-satisfying method
(Sharp Min--Max + ALM, HM 67.11) by 5.7 points.

Although BLUR attains a similar HM (71.40), its Forget ROUGE is 0.28.
Our reproduction matches BLUR's reported Forget Quality, but answer-level
evaluation still reveals substantial forget-set leakage.
We therefore include BLUR for completeness, while interpreting it as outside
the strict answer-level forgetting regime targeted by SAUL.

\paragraph{Effect of ALM on Utility and Generalization.}
Figures~\ref{fig:hm_forget_tradeoff} and~\ref{fig:motivation_experiment} show that adding ALM to baseline methods improves both model utility and GPT-based evaluation metrics.
While improvements on the \textit{Retain} set are moderate, the gains are larger on the \textit{Real Authors} subset, which evaluates generalization to neighborhood data.
This pattern aligns with SAUL's design principle of forgetting only as much as necessary: once the criterion is met, ALM deactivates forget-side pressure and avoids unnecessary updates that erode semantically related knowledge in neighborhood subsets, whereas open-ended forgetting objectives continue applying forget-side gradients and degrade such data more aggressively.

We further evaluate this behavior under paraphrased questions, where forget queries preserve the original semantics but vary in lexical and structural form.
As shown in Table~\ref{tab:forget1_qpara}, SAUL remains effective under this query perturbation, suggesting that its forgetting behavior is not tied to the surface form of the original questions.
Additional paraphrased-question results for other forgetting ratios and larger backbones are provided in Appendix~\ref{app:additional_results}, where the same overall trend holds.

\paragraph{Efficacy and Constraint Satisfaction.} 
As shown in Table~\ref{tab:forget1}, SAUL achieves strong overall performance while satisfying the forgetting constraint.
Adding ALM improves model utility over the corresponding non-ALM variants, indicating that the constraint-based formulation can enforce sufficient forgetting without excess retention degradation.

This trend holds beyond the main setting.
In Appendix~\ref{app:additional_results}, we report additional results for 5\% and 10\% forget ratios, paraphrased-question evaluation, and the larger LLaMA-3.2-3B backbone.
Across these settings, SAUL remains competitive and often achieves the best harmonic mean, showing that the proposed ALM--SAM formulation generalizes across forgetting ratios, query perturbations, and model scales.

\paragraph{Ablation Study}
\label{sec:ablation}
To separate the contributions of sharpness-aware optimization and the augmented Lagrangian formulation, we evaluate three SAUL variants: (i) removing ALM while keeping SAM, (ii) removing SAM while keeping ALM, and (iii) removing both. In full SAUL, ALM is applied to the sharpness-aware forget loss evaluated at the local perturbation most favorable to recovering the forgotten knowledge, thereby enforcing the forgetting constraint over a local weight-space neighborhood rather than only at the current parameters. In contrast, the baseline ``+\,ALM'' variants and the SAUL ``w/o SAM'' variant apply ALM to the clean forget loss, allowing us to isolate the effect of constraint enforcement on sharpness-aware and clean forget measures.

\subsection{Results on WMDP}\label{sec:exp_wmdp}

\begin{table}[t]
\centering
\small
\setlength{\tabcolsep}{4pt}
\renewcommand{\arraystretch}{1.0}
\caption{Unlearning performance on WMDP using \texttt{Zephyr-7B-$\beta$}. 
WMDP-Bio and WMDP-Cyber measure residual hazardous knowledge on the forget 
domains (lower is better), while MMLU measures general utility preservation 
(higher is better).}
\label{tab:wmdp_main}
\begin{adjustbox}{max width=\columnwidth}
\begin{tabular}{@{}lccc@{}}
\toprule
Method & Bio$\downarrow$ & Cyber$\downarrow$ & MMLU$\uparrow$ \\
\midrule
Original 
& 0.637 
& 0.440 
& 0.581 \\
\midrule
Sharp Min--Max
& \scorebf{0.260}{0.012}
& \score{0.260}{0.004}
& \score{0.535}{0.002} \\
Relearning-resilient
& \score{0.262}{0.000}
& \score{0.262}{0.000}
& \score{0.414}{0.000} \\
BLUR
& \score{0.262}{0.010}
& \score{0.253}{0.009}
& \score{0.540}{0.004} \\
\midrule
\textbf{SAUL}
& \score{0.268}{0.012}
& \scorebf{0.251}{0.010}
& \scorebf{0.542}{0.003} \\
\bottomrule
\end{tabular}
\end{adjustbox}
\end{table}

Table~\ref{tab:wmdp_main} reports results on WMDP. All compared methods are tuned to achieve comparable forgetting, with the average accuracy over WMDP-Bio and WMDP-Cyber kept near the random-chance level of $0.25$ for four-way multiple-choice questions. The averaged WMDP accuracies differ by less than $0.005$ across methods, indicating comparable benchmark-level forgetting of hazardous knowledge. Under this matched-forgetting condition, SAUL achieves an MMLU accuracy of $0.542\pm0.003$, comparable to BLUR's $0.540\pm0.004$. Sharp Min--Max achieves $0.535\pm0.002$, while Relearning-resilient exhibits a substantially lower MMLU accuracy of $0.414$. The largest utility gap is observed relative to Relearning-resilient, suggesting that aggressive forget-side robustness can incur substantial utility degradation at the evaluated operating point. Because the methods differ in multiple optimization components, this gap cannot be attributed solely to the augmented-Lagrangian controller.

\subsection{Results on MUSE Books}\label{sec:exp_muse}

\begin{table}[H]
\centering
\small
\setlength{\tabcolsep}{4pt}
\renewcommand{\arraystretch}{1.0}
\caption{Unlearning performance on MUSE Books using \texttt{Llama-2-7B}.
VerbMem and KnowMem on $\mathcal{D}_f$ measure residual memorization on the
forget set (lower is better); PrivLeak measures privacy leakage (closer to
$0$ is better); KnowMem on $\mathcal{D}_r$ measures retain-side knowledge
preservation (higher is better). BLUR and SAUL are tuned to achieve
matched knowledge forgetting (KnowMem $\mathcal{D}_f \approx 5$); other
baselines are reported at their default operating points.}
\label{tab:muse_books_main}
\begin{adjustbox}{max width=\columnwidth}
\begin{tabular}{@{}lcccc@{}}
\toprule
Method
& \makecell[c]{VerbMem\\$\downarrow$}
& \makecell[c]{KnowMem\\$\mathcal{D}_f \downarrow$}
& \makecell[c]{PrivLeak\\$\rightarrow 0$}
& \makecell[c]{KnowMem\\$\mathcal{D}_r \uparrow$} \\
\midrule
Original
& 99.8 & 59.4 & -57.5 & 66.9 \\
Retrain
& 14.3 & 28.9 & 0.0 & 74.5 \\
\midrule
Sharp Min--Max
& 0.00 $\pm$ 0.00
& 5.37 $\pm$ 1.29
& -33.53 $\pm$ 0.86
& 39.76 $\pm$ 1.82 \\
Relearning-resilient
& 0.03 $\pm$ 0.06
& 5.67 $\pm$ 0.77
& -2.13 $\pm$ 3.89
& 39.03 $\pm$ 1.03 \\
BLUR
& 0.00 $\pm$ 0.00
& 3.09 $\pm$ 4.97
& -30.77 $\pm$ 5.41
& \textbf{48.40 $\pm$ 1.50} \\
\midrule
\textbf{SAUL}
& \textbf{0.00 $\pm$ 0.00}
& \textbf{0.77 $\pm$ 1.34}
& \textbf{-16.66 $\pm$ 2.90}
& 48.25 $\pm$ 5.82 \\
\bottomrule
\end{tabular}
\end{adjustbox}
\end{table}

Table~\ref{tab:muse_books_main} reports results on MUSE Books, which evaluates the removal of verbatim and factual memorization from the Harry Potter corpus. PrivLeak measures distributional similarity to the retrained reference model, with values closer to $0$ indicating better alignment with the retrained model.

SAUL achieves the strongest overall forgetting among the compared methods, with VerbMem $=0.00\pm0.00$ and KnowMem on $\mathcal{D}_f=0.77\pm1.34$. Despite attaining a lower forget-set KnowMem than BLUR, SAUL preserves a similar mean level of retain-side knowledge ($48.25\pm5.82$ vs.\ $48.40\pm1.50$) and obtains a PrivLeak score closer to zero ($-16.66\pm2.90$ vs.\ $-30.77\pm5.41$).

The remaining baselines exhibit different trade-offs. Relearning-resilient achieves the PrivLeak score closest to the retrained reference ($-2.13\pm3.89$), but preserves substantially less retain-side knowledge than SAUL ($39.03\pm1.03$ vs.\ $48.25\pm5.82$). Sharp Min--Max similarly incurs greater retain-side degradation and obtains a PrivLeak score farther from zero than SAUL. Additional results on the MUSE News benchmark are provided in Appendix~\ref{app:muse_additional}.

\section{Conclusion} 
\label{sec:conc}

We propose \emph{Sharpness-Aware Augmented Lagrangian Unlearning (SAUL)}, which formulates LLM unlearning as constrained optimization under the principle of ``\emph{forget enough, but no more than necessary.}''
SAUL uses an augmented-Lagrangian controller to reduce forget-side pressure once a prescribed forgetting criterion is met, and combines it with sharpness-aware updates and role-separated optimizer states for stable unlearning.
Experiments on ToFU, WMDP, and MUSE Books show that SAUL improves the forgetting--utility trade-off under matched forgetting criteria, preserving retain-side utility, neighborhood-data performance, and general capability while avoiding excessive forgetting.
We further show that the augmented-Lagrangian controller can serve as a method-agnostic drop-in modifier for representative sharpness- and perturbation-based baselines, highlighting the broader value of explicit forgetting control in LLM unlearning.
\section{Limitations}
\label{sec:limitations}

SAUL requires selecting a forgetting threshold that specifies when the target knowledge is sufficiently removed.
Although this threshold provides an interpretable interface for controlling the forgetting--utility trade-off, its optimal value may depend on the dataset, model scale, forget ratio, and evaluation metric.
Developing automatic or theoretically grounded threshold-selection procedures remains an important direction for future work.

Our evaluation focuses on benchmark-level unlearning across ToFU, WMDP, and MUSE.
While we include paraphrased-question evaluation to test robustness to surface-form changes, we do not provide formal guarantees against all adversarial prompting, relearning, or recovery attacks.
Future work should study stronger adaptive attacks and broader forms of post-unlearning robustness.

Finally, SAUL introduces additional computation from sharpness-aware perturbation steps and augmented-Lagrangian updates.
Although the overhead is moderate in our experiments, scaling the method to substantially larger models or more complex unlearning targets may require further memory- and compute-efficient implementations; further details are provided in Appendix~\ref{app:computational_overhead}.
\section*{Acknowledgements}

This work was supported by Institute of Information \& communications Technology Planning \& Evaluation (IITP) and the National Research Foundation of Korea (NRF) grant funded by the Korea government (MSIT) (RS-2019-II191906, Artificial Intelligence Graduate School Program (POSTECH) (5\%); RS-2024-00457882, National AI Research Lab Project (10\%); No. RS-2024-00509258 and No. RS-2024-00469482, Global AI Frontier Lab (40\%); RS-2025-00560062 (45\%)).

\bibliography{llm,sml,tml-lab}

\appendix
\onecolumn

\newcommand{\kwwrong}[1]{\textbf{\textcolor{red!90!black}{#1}}}
\newcommand{\kwright}[1]{\textbf{\textcolor{green!60!black}{#1}}}

\section{Baselines}
\label{app:baselines}

We compare SAUL against standard optimization baselines, recent LLM unlearning methods, and ALM-augmented variants of representative sharpness-based baselines.
All methods are evaluated under the same backbone, dataset split, training budget, and evaluation metrics.

\subsection{Standard Optimization Baselines}

\paragraph{Single AdamW.}
Single AdamW optimizes the unlearning objective with a single AdamW optimizer state shared across retain- and forget-side updates.
This baseline serves as the simplest reference for assessing whether the proposed role-separated optimizer design improves stability beyond conventional adaptive optimization.

\paragraph{Dual AdamW.}
Dual AdamW maintains separate AdamW optimizer states for retain- and forget-side updates while sharing the same model parameters.
Because retain and forget objectives can induce gradients with different semantics and scales, separating their adaptive statistics can reduce optimizer-state interference.
This baseline isolates the effect of role-separated optimizer states without the full constrained sharpness-aware formulation.

\subsection{Recent LLM Unlearning Baselines}

\paragraph{Relearning-resilient Unlearning.}
Relearning-resilient Unlearning improves resistance to relearning-style recovery by applying sharpness-aware optimization to the forget-side objective~\cite{fan2025relearning}.
Its objective robustifies forgetting through an inner maximization over local weight perturbations, while the retain side enters as a weighted regularization term.
This baseline is relevant because it directly tests whether forget-side sharpness alone is sufficient, or whether explicit forgetting control and retain-side sharpness-aware updates provide additional benefits.

\paragraph{Sharp Min--Max.}
Sharp Min--Max applies different sharpness-aware dynamics to the retain and forget sides~\cite{tang2025sharpness_aware_mu}.
The retain side minimizes worst-case retain loss to encourage flatness and preserve utility, while the forget side promotes sharpness-increasing dynamics to amplify forgetting.
We use the no-masking variant in our experiments.
This baseline is closely related to SAUL because it already separates retain- and forget-side perturbation roles, but it does not adaptively deactivate forget-side pressure based on a prescribed forgetting threshold.


\paragraph{BLUR.}
BLUR is a bi-level optimization method for LLM unlearning~\cite{reisizadeh2025blurbileveloptimizationapproach}.
It is closely related to SAUL because it also departs from a simple weighted-sum formulation.
However, while BLUR structures unlearning through a bi-level optimization problem, SAUL formulates forgetting as an explicit constraint and optimizes retain-side utility subject to satisfying a target forgetting level.
This comparison evaluates the difference between bi-level trade-off optimization and forget-constrained satisficing.

\paragraph{Primal-Dual Unlearning (PDU).}
PDU is a constrained optimization method for LLM unlearning that minimizes a forget-side objective subject to an explicit retain-side utility constraint~\cite{entesari2025constrainedentropicunlearningprimaldual}. It is closely related to SAUL because both methods use primal--dual optimization to control the forgetting--utility trade-off through an explicit constraint. However, PDU constrains retain-side degradation while leaving forgetting as the primary optimization objective, whereas SAUL constrains the forget side and minimizes retain loss subject to achieving a prescribed forgetting level. This comparison evaluates the difference between retain-constrained forgetting and forget-constrained utility preservation.

\subsection{ALM-Augmented Sharpness-Based Variants}
\label{app:alm_variants}
To test whether the augmented-Lagrangian controller is useful beyond SAUL itself, we also apply it as a drop-in modifier to representative sharpness- and perturbation-based baselines.
Given a baseline objective $\mathcal{J}(\theta)$ and a forgetting measure $F(\theta)$ where larger values indicate stronger forgetting, we impose
\begin{equation*}
\min_{\theta}\ \mathcal{J}(\theta)
\quad \text{s.t.}\quad
F(\theta)\ge \alpha .
\end{equation*}
The projected multiplier update is
\begin{equation*}
\lambda^+ \leftarrow \big[\lambda+\mu(\alpha-F(\theta))\big]_+ .
\end{equation*}
The multiplier increases forget-side pressure when the constraint is violated and reduces it as the constraint remains satisfied.

\begin{table*}[t]
\centering
\small
\setlength{\tabcolsep}{5pt}
\renewcommand{\arraystretch}{1.18}
\begin{tabularx}{\textwidth}{
  @{}
  >{\raggedright\arraybackslash}p{3.2cm}
  >{\centering\arraybackslash}p{7.4cm}
  >{\raggedright\arraybackslash}X
  @{}
}
\toprule
\textbf{Method}
& \textbf{Core objective $\mathcal{J}(\theta)$}
& \makecell[l]{\textbf{Where sharpness/}\\\textbf{perturbation enters}} \\
\midrule

\makecell[l]{\textbf{Relearning-resilient}\\\textbf{baseline}}
&
\vspace{1.8em}
\objcell{
\begin{aligned}
\min_{\theta}\ 
\Big[
  \max_{\|\delta\|_{p}\le\rho}
  \mathcal{L}^{\mathrm{NPO}}_{f}(\theta+\delta)
  + \lambda\,\mathcal{L}_{r}(\theta)
\Big]
\end{aligned}
}
&
\vspace{-1.6em}
\textbf{Forget-side perturbation:}
the inner maximization over weight perturbations is applied to the forget-side objective, while the retain side remains an unperturbed weighted term. \\
\addlinespace[0.8ex]
\midrule

\makecell[l]{\textbf{Sharp Min--Max}\\\textbf{(no masking)}}
&
\objcell{
\begin{aligned}
\text{Retain:}\quad
& \min_{\theta}
  \max_{\|\delta_r\|\le\rho_r}
  \mathcal{L}_{r}(\theta+\delta_r)
\\[0.4ex]
\text{Forget:}\quad
& \min_{\theta}
  \Big\{
  \mathcal{L}_{f}(\theta)
  -
  \big[
    \max_{\|\delta_f\|\le\rho_f}
    \mathcal{L}_{f}(\theta+\delta_f)
    -
    \mathcal{L}_{f}(\theta)
  \big]
  \Big\}
\end{aligned}
}
&
\vspace{-1.6em}
\textbf{Two-role perturbation dynamics:}
the retain side minimizes worst-case retain loss for utility preservation, while the forget side promotes sharpness-increasing dynamics to amplify forgetting. \\
\bottomrule
\end{tabularx}
\caption{Summary of representative sharpness- and perturbation-based unlearning objectives considered for ALM augmentation. $\mathcal{L}_r$ and $\mathcal{L}_f$ denote retain and forget losses; $\rho,\rho_r,\rho_f$ are weight-space perturbation radii; $\lambda$ denotes a trade-off coefficient in the original baseline objective.}
\label{tab:sam_family_objectives}
\end{table*}

\paragraph{ALM-augmented Relearning-resilient baseline.}
For Relearning-resilient Unlearning~\cite{fan2025relearning}, the original baseline applies sharpness-aware optimization to the forget-side objective while using $\lambda\mathcal{L}_r(\theta)$ as a retain-side weighted term.
We define the forgetting measure as
\begin{equation*}
F_{\mathrm{relearn}}(\theta)
\triangleq
\max_{\|\delta\|\le\rho}\ \mathcal{L}^{\mathrm{NPO}}_{f}(\theta+\delta).
\end{equation*}
Enforcing $F_{\mathrm{relearn}}(\theta)\ge \alpha$ requires the forget-side objective to remain large under local weight perturbations.
The fixed trade-off coefficient is then replaced by the adaptive multiplier induced by the augmented-Lagrangian controller.

\paragraph{ALM-augmented Sharp Min--Max.}
For Sharp Min--Max~\cite{tang2025sharpness_aware_mu}, the retain-side update minimizes worst-case retain loss, while the forget-side update promotes sharpness-increasing dynamics.
We apply ALM at the update level: the retain-side SAM update is kept intact, and the multiplier activates and scales the forget-side update when the forgetting constraint is violated.
As $F(\theta)$ remains above $\alpha$, the projected multiplier can decrease toward zero, reducing or eventually skipping the forget-side step so that optimization focuses on retain-side sharpness minimization.
Compared with these ALM-augmented baselines, SAUL combines three design choices in a single method: explicit forget-side constraint control, sharpness-aware updates on both retain and forget objectives, and role-separated optimizer states.

\paragraph{Implementation and Evaluation.}
We tune method-specific hyperparameters using a comparable validation protocol and follow the original recommended settings when available.
For all methods, we select operating points under matched forgetting criteria so that comparisons reflect utility differences at comparable levels of forgetting.
SAUL is implemented following Algorithm~\ref{alg:main}, which combines sharpness-aware updates, augmented-Lagrangian multiplier control, and separate optimizer states for the retain- and forget-side objectives.

\begin{algorithm}[ht]
\caption{Sharpness-aware Augmented Lagrangian Method (SAUL)}
\label{alg:main}
\begin{algorithmic}[1]
\REQUIRE Initial parameters $\theta$, multiplier $\lambda\ge0$, optimizer states $s_r,s_f$;
retain batch $\mathcal{B}_r$, forget batch $\mathcal{B}_f$; SAM radii $\rho_r,\rho_f$; ALM penalty $\mu>0$; threshold $\alpha$;
small $\epsilon_{\text{pert}}$.

\STATE \textbf{procedure} \textsc{SAUL}$(\theta,\lambda,s_r,s_f,\mathcal{B}_r,\mathcal{B}_f,\rho_r,\rho_f,\mu,\alpha,\epsilon_{\text{pert}})$
\FOR{$t = 1,2,\dots,T$}
    \STATE $(g_r^{\text{SAM}}, \delta^*_r) \leftarrow \textsc{Grad}_{\text{SAM}}(\theta, \mathcal{B}_r, \mathcal L_r, \rho_r, \texttt{retain})$
    \STATE $(g_f^{\text{SAM}}, \delta^*_f) \leftarrow \textsc{Grad}_{\text{SAM}}(\theta, \mathcal{B}_f, \mathcal L_f, \rho_f, \texttt{forget})$
    \STATE $c_f \leftarrow \alpha - \mathcal L_f(\theta + \delta_f^*;\mathcal{B}_f)$
    \STATE $\lambda^+ \leftarrow \big[\lambda + \mu\,c_f\big]_+$
    \IF{$\lambda^+ > 0$}
        \STATE $\tilde{g}_f \leftarrow -\lambda^+\, g_f^{\text{SAM}}$
        \STATE $(\theta, s_f) \leftarrow \textsc{AdamW}(\theta, s_f, \tilde{g}_f)$
    \ENDIF
    \STATE $(\theta, s_r) \leftarrow \textsc{AdamW}(\theta, s_r, g_r^{\text{SAM}})$
    \STATE $\lambda \leftarrow \lambda^+$
\ENDFOR
\RETURN $\theta, \lambda, s_r, s_f$

\medskip

\STATE \textbf{procedure} \textsc{Grad$_{\text{SAM}}$}$(\theta, \mathcal{B}, \mathcal L, \rho, \texttt{mode})$
\STATE \hspace{1em} $g \leftarrow \nabla_\theta \mathcal L(\theta; \mathcal{B})$
\STATE \hspace{1em} \textbf{if} $\texttt{mode} = \texttt{retain}$ \textbf{then} $\delta^{\star} \leftarrow \rho \, \dfrac{g}{\|g\|_2 + \epsilon_{\text{pert}}}$
\STATE \hspace{1em} \textbf{else} \hspace{1.35em} $\delta^{\star} \leftarrow -\rho \, \dfrac{g}{\|g\|_2 + \epsilon_{\text{pert}}}$
\STATE \hspace{1em} $g^{\text{SAM}} \leftarrow \nabla_\theta \mathcal L(\theta + \delta^{\star}; \mathcal{B})$
\STATE \hspace{1em} \textbf{return} $(g^{\text{SAM}}, \delta^{\star})$
\STATE \textbf{end procedure}

\end{algorithmic}
\end{algorithm}

\clearpage
\section{Experimental Setup}
\label{app:experimental_setup}

\subsection{Training Configuration}
\label{app:training_config}

For TOFU, models are trained for 10 epochs with a batch size of 8.
For WMDP, we follow the RMU training setup and train for one epoch over at most 150 mini-batches with a batch size of 4, using two GPUs to host the updated and frozen models.
For MUSE-Books, we train for 10 epochs with a per-device batch size of 1, using gradient checkpointing when needed for memory efficiency.
Our method uses AdamW-style parameter updates, maintaining separate optimizer states for the forget and retain objectives unless otherwise specified by a baseline.
Experiments are conducted on NVIDIA RTX 6000/A6000-class GPUs for TOFU and WMDP, and on H200 and RTX Pro 6000-class GPUs for MUSE-Books.
For TOFU and WMDP, we report mean and standard deviation over five independent runs.
For MUSE-Books, we report results from a single run due to the substantially higher computational cost of 7B-scale memorization evaluation.

\subsection{Hyperparameter Selection}
\label{app:hyperparameter_selection} 

Hyperparameters for all methods are selected via grid search.

\begin{table}[H]
\centering
\small
\caption{Final hyperparameter configurations for the main ToFU setting with forget ratio $=1\%$.}
\label{tab:tofu_hyperparams_forget01}
\begin{tabular}{lcccc}
\toprule
Method
& Learning rate
& SAM $(\rho_f,\rho_r)$ 
& Risk $\alpha$
& $\lambda$ LR \\
\midrule

\textbf{SAUL}
& $1\times10^{-5}$
& $1\times10^{-3}$
& $10$
& $1\times10^{-3}$ \\

w/o ALM
& $1\times10^{-5}$
& $1\times10^{-3}$
& --
& -- \\

w/o SAM
& $1\times10^{-5}$
& $(-, 1\times10^{-3})$
& $30$
& $1\times10^{-4}$ \\

w/o ALM + SAM
& $1\times10^{-5}$
& $(-, 1\times10^{-3})$
& --
& -- \\

\midrule

Dual AdamW
& $1\times10^{-5}$
& --
& --
& -- \\

Dual AdamW + ALM
& $1\times10^{-5}$
& --
& $10$
& $1\times10^{-4}$ \\

Sharp Min--Max
& $3\times10^{-2}$
& $1\times10^{-4}$
& --
& -- \\

Sharp Min--Max + ALM
& $3\times10^{-2}$
& $1\times10^{-4}$
& $10$
& $1\times10^{-4}$ \\

Relearning-resilient
& $6\times10^{-2}$
& $1\times10^{-2}$
& --
& -- \\

Relearning-resilient + ALM
& $6\times10^{-2}$
& $1\times10^{-2}$
& $20$
& $1\times10^{-4}$ \\

\bottomrule
\end{tabular}
\end{table}

\begin{table}[H]
\centering
\small
\caption{Final hyperparameter configurations for the main ToFU setting with forget ratio $=5\%$.}
\label{tab:tofu_hyperparams_forget05}
\begin{tabular}{lcccc}
\toprule
Method
& Learning rate
& SAM $(\rho_f,\rho_r)$ 
& Risk $\alpha$
& $\lambda$ LR \\
\midrule

\textbf{SAUL}
& $1\times10^{-5}$
& $1\times10^{-4}$
& $10$
& $1\times10^{-4}$ \\

w/o ALM
& $1\times10^{-5}$
& $(1\times10^{-3},\,1\times10^{-4})$
& --
& -- \\

w/o SAM
& $1\times10^{-5}$
& $(-, 1\times10^{-2})$
& $10$
& $1\times10^{-4}$ \\

w/o ALM + SAM
& $1\times10^{-5}$
& $(-, 1\times10^{-2})$
& --
& -- \\

\midrule

Dual AdamW
& $1\times10^{-8}$
& --
& --
& -- \\

Dual AdamW + ALM
& $1\times10^{-5}$
& --
& $10$
& $1\times10^{-4}$ \\

Sharp Min--Max
& $1\times10^{-1}$
& $1\times10^{-3}$
& --
& -- \\

Sharp Min--Max + ALM
& $1\times10^{-1}$
& $1\times10^{-3}$
& $20$
& $1\times10^{-4}$ \\

\bottomrule
\end{tabular}
\end{table}

\begin{table}[H]
\centering
\small
\caption{Final hyperparameter configurations for the main ToFU setting with forget ratio $=10\%$.}
\label{tab:tofu_hyperparams_forget10}
\begin{tabular}{lcccc}
\toprule
Method
& Learning rate
& SAM $(\rho_f,\rho_r)$ 
& Risk $\alpha$
& $\lambda$ LR \\
\midrule

\textbf{SAUL}
& $1\times10^{-5}$
& $1\times10^{-2}$
& $10$
& $1\times10^{-4}$ \\

w/o ALM
& $1\times10^{-5}$
& $1\times10^{-2}$
& --
& -- \\

w/o SAM
& $1\times10^{-5}$
& $(- , 1\times10^{-2})$
& $10$
& $1\times10^{-4}$ \\

w/o ALM + SAM
& $1\times10^{-5}$
& $(-, 1\times10^{-2})$
& --
& -- \\

\midrule

Dual AdamW
& $1\times10^{-5}$
& --
& --
& -- \\

Dual AdamW + ALM
& $1\times10^{-5}$
& --
& $12$
& $1\times10^{-4}$ \\

Sharp Min--Max
& $1\times10^{-2}$
& $(1\times10^{-4},\,1\times10^{-3})$
& --
& -- \\

Sharp Min--Max + ALM
& $1\times10^{-2}$
& $1\times10^{-4}$
& $20$
& $1\times10^{-4}$ \\

\bottomrule
\end{tabular}
\end{table}

For TOFU, we tune the forget risk threshold $\alpha$ over $[5, 30]$, the SAM perturbation radius $\rho$ over $[10^{-5}, 10^{-2}]$, AdamW learning rates over $[10^{-5}, 5\times10^{-5}]$, and SGD learning rates over $[10^{-3}, 10^{-1}]$.
We select configurations that reduce Forget ROUGE to around $0.03$ while preserving retain performance.
The best-found hyperparameter values for the ToFU settings with forget ratios of $1\%$, $5\%$, and $10\%$ are reported in Tables~\ref{tab:tofu_hyperparams_forget01}, \ref{tab:tofu_hyperparams_forget05}, and \ref{tab:tofu_hyperparams_forget10}, respectively.
For BLUR, we follow the hyperparameters reported in the original BLUR paper.

\begin{table}[H]
\centering
\small
\caption{Final hyperparameter configurations for WMDP. The RMU steering coefficient is fixed to $7.5$ except for Sharp Min--Max, which uses $6.5$. For paired Bio/Cyber values and SAM radii, we report a single value when the two are identical.}
\label{tab:wmdp_hyperparams}
\resizebox{\linewidth}{!}{
\begin{tabular}{lccccccc}
\toprule
Method 
& RMU weight 
& Steering 
& Forget LR 
& Retain LR 
& Joint LR 
& SAM $(\rho_f,\rho_r)$ 
& Risk $\alpha$ \\
\midrule

Sharp Min--Max
& $1200$ 
& $6.5$ 
& $5\times10^{-5}$ 
& $5\times10^{-5}$ 
& -- 
& $3\times10^{-4}$ 
& -- \\

Relearning-resilient Unlearning
& $1200$ 
& $7.5$ 
& -- 
& -- 
& $1.25\times10^{-5}$ 
& $5\times10^{-6}$ 
& -- \\

BLUR 
& $800$ 
& $7.5$ 
& -- 
& -- 
& $6.5\times10^{-5}$ 
& -- 
& -- \\

\textbf{SAUL}
& $800$ 
& $7.5$ 
& $6\times10^{-6}$ 
& $4\times10^{-5}$ 
& -- 
& $1\times10^{-4}$ 
& $1.5\times10^{-3}$ \\

\bottomrule
\end{tabular}
}
\end{table}

For WMDP, we tune the remaining hyperparameters around the best-performing regions. The RMU retain coefficient is tuned over $[750,1200]$ depending on the method. For AdamW-based variants, we tune learning rates over
$[5\times10^{-6}, 7\times10^{-5}]$, using a single joint learning rate for BLUR and Relearning-resilient Unlearning and separate forget-side and retain-side learning rates for Sharp Min--Max and SAUL. For constrained variants, we tune the forget risk threshold $\alpha$ over $[10^{-3}, 1.5\times10^{-3}]$ and the SAM perturbation radii $(\rho_f,\rho_r)$ over $[5\times10^{-6}, 3\times10^{-4}]$, while fixing $\beta=0.1$ and the Lagrange multiplier learning rate to $10^{-3}$. Final configurations are selected to reduce both WMDP-Bio and WMDP-Cyber accuracy to near-chance levels while preserving MMLU performance.

\begin{table}[H]
\centering
\small
\caption{Final hyperparameter configurations for MUSE-Books. For paired SAM radii, we report a single value when the two are identical.}
\label{tab:muse_books_hyperparams}
\resizebox{\linewidth}{!}{
\begin{tabular}{lcccccc}
\toprule
Method
& Forget LR
& Retain LR
& Base LR
& $\beta$
& SAM $(\rho_f,\rho_r)$
& Risk $\alpha$ \\
\midrule

Sharp Min--Max
& --
& --
& $1\times10^{-5}$
& --
& --
& -- \\

Relearning-resilient Unlearning
& --
& --
& $2.15\times10^{-6}$
& $0.1$
& $5\times10^{-3}$
& -- \\

BLUR
& --
& --
& $2.4\times10^{-6}$
& $0.4$
& --
& -- \\

\textbf{SAUL}
& $2\times10^{-5}$
& $6\times10^{-5}$
& --
& $0.1$
& $5\times10^{-3}$
& $55$ \\

\bottomrule
\end{tabular}
}
\end{table}

For MUSE-Books, we tune learning rates over method-specific grids within
$[1.0\times10^{-6}, 6.0\times10^{-5}]$, using separate forget- and
retain-side learning rates when applicable. For Sharp Min--Max, we use
$\gamma=0.2$ and forget scale $1.0$. For relearning-resilient unlearning,
we set the resilient objective coefficient to $2.0$ and the NPO coefficient
to $0.75$. For BLUR, we use $\gamma=1.0$. For SAUL, we tune the forget-risk
threshold $\alpha$ on the scale of the corresponding forget objective and
use $\lambda_0=1.0$ with Lagrange multiplier learning rate
$3\times10^{-4}$. Final configurations are selected based on forgetting,
privacy leakage, verbatim memorization, and retain-side stability under the
matched-forgetting condition $\mathrm{KnowMem}(\mathcal{D}_f)\approx 5$.

\begin{table}[H]
\centering
\small
\caption{Final hyperparameter configurations for MUSE-News. For paired SAM radii, we report a single value when the two are identical.}
\label{tab:muse_news_hyperparams}
\resizebox{\linewidth}{!}{
\begin{tabular}{lcccccc}
\toprule
Method
& Forget LR
& Retain LR
& Base LR
& $\beta$
& SAM $(\rho_f,\rho_r)$
& Risk $\alpha$ \\
\midrule

Sharp Min--Max
& --
& --
& $2.66\times10^{-4}$
& --
& --
& -- \\

Relearning-resilient Unlearning
& --
& --
& $7\times10^{-5}$
& $0.2$
& $5\times10^{-3}$
& -- \\

BLUR
& --
& --
& $6.5\times10^{-5}$
& $0.4$
& --
& -- \\

\textbf{SAUL}
& $2.25\times10^{-5}$
& $6\times10^{-5}$
& --
& $0.1$
& $5\times10^{-3}$
& $40$ \\

\bottomrule
\end{tabular}
}
\end{table}

For MUSE-News, we tune learning rates over method-specific grids, using
separate forget- and retain-side learning rates when applicable. For Sharp
Min--Max, we use $\gamma=0.05$ and forget scale $20.0$. For
relearning-resilient unlearning, we set the resilient objective coefficient
to $5.0$ and the NPO coefficient to $2.0$. For BLUR, we use $\gamma=1.0$.
For SAUL, we tune the forget-risk threshold $\alpha$ on the scale of the
corresponding forget objective and use $\lambda_0=0.45$ with Lagrange
multiplier learning rate $7.5\times10^{-5}$. Final configurations are
selected based on forgetting, privacy leakage, verbatim memorization, and
retain-side stability under the matched-forgetting condition
$\mathrm{KnowMem}(\mathcal{D}_f)\approx 30$.

\subsection{Unlearning Optimization Details}
\label{app:unlearning_optimization}

For SAUL, SAM is applied separately to the retain and forget objectives.
The retain-side update promotes utility preservation under local parameter
perturbations, whereas the forget-side update enforces robust forgetting
within the augmented Lagrangian formulation. Unless otherwise specified,
the same SAM perturbation radius and optimizer learning rate are used for
both update pathways.

\subsection{GPT-based Evaluation Protocol}
\label{app:gpt_eval_protocol}
We use GPT-5.1 as the evaluator for GPT-based semantic evaluation, assessing whether model-generated answers are semantically equivalent to the reference answers. The same prompt template is used across all GPT-based evaluations in ToFU.

\noindent\textbf{Prompt Template.}
The following prompt is used for all GPT-based evaluations:

\noindent\fbox{%
\parbox{\linewidth}{%
\textbf{You are an expert in Natural Language Inference.}  
Your task is to determine whether the following two responses convey the same meaning.
The first and second response are each a single sentence or word.

Please provide an answer with a score of \texttt{0} or \texttt{1}, where:  
\texttt{0} = contradiction  
\texttt{1} = entailment

\textbf{Question:} \{question\}

\textbf{Response 1 (sentence):} \{generation\}

\textbf{Response 2 (sentence/word):} \{ground\_truth\}

Provide your score only as a number.

\textbf{Score:}
}}

\subsection{Qualitative Analysis of Generated Responses}
\label{app:qualitative_results}

The following examples are drawn from the Forget subset.
Table~\ref{tab:qualitative_forget_example} presents representative examples of model responses on the Forget set, while Table~\ref{tab:qualitative_retain_example} shows corresponding examples from the Retain set to illustrate knowledge preservation after unlearning.

\begin{table}[ht]
\centering
\small
\setlength{\tabcolsep}{4pt}
\caption{Qualitative example from the ToFU Forget set.}
\renewcommand{\arraystretch}{1.15}
\begin{tabular}{p{\linewidth}}
\toprule
\textbf{Forget Example (ToFU)} \\
\midrule
\textbf{Question:} {What is the full name of the author born in Kuwait City, Kuwait on 08/09/1956?} \\[0.2em]
\textbf{Ground Truth:} {The full name of the fictitious author born in Kuwait City, Kuwait on the 8th of September, 1956 is \textbf{Basil Mahfouz Al-Kuwaiti.}} \\
\midrule
\textbf{Answers} \\[0.2em]  
\textbf{Sharpness Min-Max:} \xmark\ {The full name of the author born in Kuwait City, Kuwait on 08/09/1956 is \kwwrong{Jad Ambrose Al-Shamary}} \\[0.15em]
\textbf{Dual-AdamW:} \xmark\ {The author's full name is \kwwrong{Basma} experience experience experience, an authentic voice from experience coming experience from experience experience Kuwait about city experience experience experience experience.} \\[0.15em]
\textbf{SAUL (Ours):} \xmark\ {The author's full name is \kwwrong{Leila Al-Sabah}, a name derived from her birthplace, gender, and year of birth. }\\
\bottomrule
\end{tabular}
\label{tab:qualitative_forget_example}

\end{table}

\begin{table}[ht]
\centering
\small
\setlength{\tabcolsep}{4pt}
\caption{Qualitative example from the ToFU Retain set.}
\renewcommand{\arraystretch}{1.15}
\begin{tabular}{p{\linewidth}}
\toprule
\textbf{Retain Example (ToFU)} \\
\midrule
\textbf{Question:} {What genre does Chukwu Akabueze specialize in?} \\[0.2em]
\textbf{Ground Truth:} {Chukwu Akabueze is a specialist in the \textbf{Biography} genre} \\
\midrule
\textbf{Answers} \\[0.2em]  
\textbf{Sharpness Min-Max:} \xmark\ {Chukwu Akabueze specializes in the genre of \kwwrong{Literature.}} \\[0.15em]
\textbf{Dual-AdamW:} \xmark\ {Chukwu Akabueze primarily specializes in the genre of \kwwrong{Historical Fiction}} \\[0.15em]
\textbf{SAUL (Ours):} \cmark\ {Chukwu Akabueze specializes in the genre of \kwright{Biography}} \\
\bottomrule
\end{tabular}
\label{tab:qualitative_retain_example}
\end{table}

\subsection{Artifact Use and Licenses}

\label{app:artifact_use}

We use publicly available research artifacts for benchmark-level evaluation, including the ToFU, WMDP, and MUSE Books benchmarks, as well as the model backbones described in Section~\ref{sec:exp_setup}. ToFU and WMDP are released under the MIT license. MUSE Books and the Llama model backbones are used according to the licenses, access conditions, and acceptable-use terms specified in their official releases. Zephyr-7B-$\beta$ is released under MIT.

These artifacts are used only for research evaluation of machine unlearning methods, consistent with their benchmark and model-evaluation purposes. We do not redistribute the original datasets, model checkpoints, or benchmark resources as part of this submission. Users should consult the original artifact releases for their applicable licenses, access conditions, and terms of use.

For the GPT-paraphrased ToFU questions constructed in this work, the intended use is evaluation of robustness to query rephrasing in machine unlearning research; these paraphrases are derived from benchmark questions and should be used only in research contexts compatible with the original benchmark conditions.

\clearpage

\section{Paraphrased ToFU Construction and Validation}
\label{app:paraphrased_tofu}

We construct a paraphrased version of the ToFU evaluation questions to evaluate whether unlearning behavior remains stable under query rephrasing. The paraphrased questions are used only for evaluation and are not included in unlearning training.

\subsection{Paraphrase Generation}
\label{app:paraphrase_generation}

For each original ToFU question, we generate a single paraphrased question while keeping the corresponding ground-truth answer unchanged. The generation prompt is designed to preserve the semantic target of the original question while encouraging lexical and syntactic diversity. Specifically, we use the following instruction:

\noindent\fbox{%
\parbox{\dimexpr\linewidth-2\fboxsep-2\fboxrule\relax}{%
\small
\textbf{You are generating paraphrased questions for robustness evaluation.}

\vspace{0.3em}
\textbf{Follow ALL rules strictly:}

1. Preserve the exact meaning. \\
2. The correct answer must remain identical. \\
3. Use substantially different wording. \\
4. Change the sentence structure. \\
5. Avoid copying more than 3 consecutive words. \\
6. Minimize token overlap as much as possible. \\
7. Do NOT introduce or remove information. \\
8. Keep the same question type (who/what/when/etc). \\
9. Output ONLY the paraphrased question. \\

\vspace{0.3em}
Do not replace domain-specific concepts (e.g., genre, profession, nationality) with related but distinct concepts. Preserve the exact semantic target of the question.
}}

\subsection{Paraphrase Quality Validation}
\label{app:paraphrase_validation}

We validate the generated paraphrases using both lexical-overlap and semantic-preservation metrics. BLEU is computed by treating the original question as the reference and the paraphrased question as the hypothesis, using \texttt{sacrebleu.corpus\_bleu} divided by $100$. ROUGE-L is computed as ROUGE-L F1, which measures overlap based on the longest common subsequence. For semantic preservation, we compute the cosine similarity between SBERT embeddings from \texttt{sentence-transformers/all-MiniLM-L6-v2}. Thus, high-quality paraphrases should have low BLEU and ROUGE-L F1, indicating reduced lexical and structural overlap, while maintaining high semantic similarity.

\begin{table}[ht]
\centering
\small
\setlength{\tabcolsep}{5pt}
\renewcommand{\arraystretch}{1.05}
\caption{Validation statistics for the GPT-paraphrased ToFU questions. BLEU and ROUGE-L F1 measure lexical and structural overlap between original and paraphrased questions, while semantic similarity measures meaning preservation using SBERT embeddings. Values are reported as mean $\pm$ standard deviation over question pairs.}
\begin{tabular}{lcccc}
\toprule
Split & $N$ & BLEU $\downarrow$ & ROUGE-L F1 $\downarrow$ & Semantic Sim. $\uparrow$ \\
\midrule
Forget & 400 & $0.208 \pm 0.141$ & $0.517 \pm 0.130$ & $0.929 \pm 0.040$ \\
Retain & 400 & $0.175 \pm 0.107$ & $0.474 \pm 0.125$ & $0.928 \pm 0.035$ \\
Real Authors & 100 & $0.274 \pm 0.136$ & $0.511 \pm 0.114$ & $0.944 \pm 0.023$ \\
World Facts & 117 & $0.242 \pm 0.126$ & $0.517 \pm 0.132$ & $0.916 \pm 0.052$ \\
\midrule
Overall & 1017 & $0.205 \pm 0.130$ & $0.499 \pm 0.128$ & $0.929 \pm 0.039$ \\
\bottomrule
\end{tabular}
\label{tab:paraphrase_validation}
\end{table}

Across all $1{,}017$ paraphrased questions, the average BLEU and ROUGE-L F1 scores are $0.205$ and $0.499$, respectively, indicating that the paraphrases substantially reduce lexical and structural overlap with the original questions. At the same time, the average semantic similarity remains high at $0.929$, suggesting that the paraphrased questions largely preserve the original semantic content. These results support the use of the paraphrased ToFU set as a controlled evaluation of robustness to query rephrasing.

\subsection{Examples}
\label{app:paraphrase_examples}

Table~\ref{tab:paraphrase_examples} shows representative examples of original and paraphrased questions.

\clearpage

\begin{table}[ht]
\centering
\small
\setlength{\tabcolsep}{4pt}
\renewcommand{\arraystretch}{1.1}
\caption{Examples of original and paraphrased ToFU questions.}
\begin{tabular}{p{0.46\linewidth}p{0.46\linewidth}}
\toprule
Original Question & Paraphrased Question \\
\midrule
What is the full name of the author born in Taipei, Taiwan on 05/11/1991 who writes in the genre of leadership?
&
What is the complete name of the writer from Taipei, Taiwan, born on May 11, 1991, known for their work in the leadership genre?
\\
\midrule
What does Hsiao Yun-Hwa identify as in terms of gender?
&
How does Hsiao Yun-Hwa classify herself regarding gender?
\\
\midrule
Where would you find the Eiffel Tower?
&
In which location is the Eiffel Tower situated?
\\
 \bottomrule
\end{tabular}
\label{tab:paraphrase_examples}
\end{table}

\clearpage

\section{Additional Unlearning Results}
\label{app:additional_results}
    
\subsection{Results on ToFU}


\begin{table*}[ht]
\centering
\small
\setlength{\tabcolsep}{3.0pt}
\renewcommand{\arraystretch}{0.95}
\caption{Unlearning performance with forget ratio = 5\% (same notation as Table ~\ref{tab:forget1}).}
\begin{tabular}{@{}l || cc|ccc|cc@{}}
\toprule
Method
& \multicolumn{2}{c|}{\textbf{Automatic}}
& \multicolumn{5}{c}{\textbf{GPT-based}} \\
\cmidrule(r){2-3}\cmidrule(l){4-8}
& MU$\uparrow$ & $F_{\text{ROUGE}}\downarrow$
& Retain$\uparrow$
& World Facts$\uparrow$
& Real Authors$\uparrow$
& HM $\uparrow$
& Forget$\downarrow$ \\
\midrule

Original Model
& 0.60 & 0.83 & 81.8 & 80.3 & 81.0 & 81.0 & 86.0 \\

\midrule

Single AdamW
& \meanpm{0.54}{0.011} & \textbf{\meanpm{0.01}{0.006}}
& \meanpm{54.80}{1.04}
& \meanpm{76.07}{1.35}
& \meanpm{62.60}{1.14}
& \meanpm{63.32}{0.59}
& \underline{\meanpm{0.40}{0.55}} \\

Dual AdamW
& \meanpm{0.56}{0.008} & \textbf{\meanpm{0.01}{0.005}}
& \meanpm{61.05}{1.15}
& \meanpm{81.03}{0.72}
& \meanpm{64.00}{2.35}
& \meanpm{67.63}{1.10}
& \underline{\meanpm{0.40}{0.89}} \\

Dual AdamW + ALM
& \underline{\meanpm{0.57}{0.004}} & \textbf{\meanpm{0.01}{0.003}}
& \meanpm{63.15}{0.60}
& \underline{\meanpm{82.74}{0.94}}
& \textbf{\meanpm{79.60}{1.67}}
& \underline{\meanpm{74.09}{0.50}}
& \meanpm{0.76}{1.05} \\

Sharp Min--Max
& \meanpm{0.53}{0.004} & \textbf{\meanpm{0.01}{0.003}}
& \meanpm{52.79}{0.53}
& \meanpm{78.46}{1.53}
& \meanpm{63.60}{1.14}
& \meanpm{63.27}{0.37}
& \underline{\meanpm{0.40}{0.42}} \\

Sharp Min--Max + ALM
& \meanpm{0.53}{0.004} & \textbf{\meanpm{0.01}{0.007}}
& \meanpm{54.70}{0.81}
& \meanpm{80.17}{0.94}
& \meanpm{71.60}{1.14}
& \meanpm{67.07}{0.45}
& \meanpm{0.70}{0.67} \\

Relearning-resilient
& \meanpm{0.02}{0.009} & \textbf{\meanpm{0.01}{0.006}}
& - & - & - & - & - \\

Relearning-resilient + ALM 
& \meanpm{0.02}{0.002} & \textbf{\meanpm{0.01}{0.003}}
& - & - & - & - & - \\

BLUR
& \meanpm{0.59}{0.014} & \meanpm{0.31}{0.142}
& \meanpm{72.70}{2.40}
& \meanpm{79.30}{2.00}
& \meanpm{75.80}{1.30}
& \meanpm{75.80}{1.60}
& \textcolor{red}{\meanpm{14.90}{1.10}} \\

PDU
& \textbf{\meanpm{0.58}{0.005}} & \meanpm{0.05}{0.011}
& \textbf{\meanpm{75.45}{1.52}}
& \textbf{\meanpm{82.91}{2.26}}
& \meanpm{71.60}{1.14}
& \textbf{\meanpm{76.37}{1.47}}
& \meanpm{2.40}{1.82} \\

\midrule

\textbf{SAUL}
& \textbf{\meanpm{0.58}{0.004}} & \underline{\meanpm{0.02}{0.002}}
& \underline{\meanpm{66.20}{1.20}}
& \meanpm{80.00}{1.97}
& \underline{\meanpm{75.40}{1.14}}
& \meanpm{73.39}{0.55}
& \textbf{\meanpm{0.12}{0.22}} \\

w/o ALM
& \meanpm{0.54}{0.004} & \textbf{\meanpm{0.01}{0.003}}
& \meanpm{61.15}{0.96}
& \meanpm{78.29}{0.97}
& \meanpm{68.80}{0.84}
& \meanpm{68.70}{0.33}
& \underline{\meanpm{0.40}{0.42}} \\

w/o SAM
& \meanpm{0.53}{0.015} & \textbf{\meanpm{0.01}{0.004}}
& \meanpm{62.40}{0.38}
& \meanpm{78.29}{0.97}
& \meanpm{62.60}{1.14}
& \meanpm{67.00}{0.49}
& \meanpm{0.50}{0.50} \\

w/o ALM + SAM
& \meanpm{0.55}{0.006} & \textbf{\meanpm{0.01}{0.006}}
& \meanpm{58.25}{0.79}
& \meanpm{79.32}{1.64}
& \meanpm{57.40}{1.52}
& \meanpm{63.55}{0.92}
& \meanpm{0.52}{0.69} \\

\bottomrule
\end{tabular}
\label{tab:forget05}
\end{table*}


\begin{table*}[ht]
\centering
\small
\setlength{\tabcolsep}{3.0pt}
\renewcommand{\arraystretch}{0.95}
\caption{Unlearning performance with forget ratio = 5\% under paraphrased question (same notation as Table ~\ref{tab:forget1_qpara}).}
\begin{tabular}{@{}l || cc|ccc|cc@{}}
\toprule
Method
& \multicolumn{2}{c|}{\textbf{Automatic}}
& \multicolumn{5}{c}{\textbf{GPT-based}} \\
\cmidrule(r){2-3}\cmidrule(l){4-8}
& MU$\uparrow$ & $F_{\text{ROUGE}}\downarrow$
& Retain$\uparrow$
& World Facts$\uparrow$
& Real Authors$\uparrow$
& HM $\uparrow$
& Forget$\downarrow$ \\
\midrule

Original Model
& 0.54 & 0.45 & 57.0 & 78.6 & 77.0 & 69.4 & 61.5 \\

\midrule

Single AdamW
& \meanpm{0.48}{0.01} & \textbf{\meanpm{0.02}{0.01}}
& \meanpm{48.70}{0.77}
& \meanpm{70.78}{4.65}
& \meanpm{38.20}{6.18}
& \meanpm{49.01}{3.58}
& \meanpm{0.80}{0.57} \\

Dual AdamW
& \underline{\meanpm{0.51}{0.02}} & \textbf{\meanpm{0.02}{0.01}}
& \meanpm{52.48}{1.65}
& \meanpm{71.10}{7.97}
& \meanpm{70.60}{8.20}
& \meanpm{63.26}{4.41}
& \textbf{\meanpm{0.10}{0.22}} \\

Dual AdamW + ALM
& \textbf{\meanpm{0.53}{0.01}} & \underline{\meanpm{0.03}{0.00}}
& \underline{\meanpm{52.72}{2.00}}
& \meanpm{75.56}{4.92}
& \underline{\meanpm{75.20}{4.97}}
& \underline{\meanpm{65.88}{3.06}}
& \meanpm{0.40}{0.65} \\

Sharp Min--Max
& \meanpm{0.43}{0.08} & \underline{\meanpm{0.03}{0.02}}
& \meanpm{42.78}{2.64}
& \meanpm{69.40}{7.22}
& \meanpm{69.00}{7.31}
& \meanpm{57.32}{4.63}
& \meanpm{3.40}{2.72} \\

Sharp Min--Max + ALM
& \meanpm{0.50}{0.02} & \meanpm{0.05}{0.02}
& \meanpm{49.18}{1.90}
& \meanpm{76.58}{1.52}
& \textbf{\meanpm{76.00}{1.41}}
& \meanpm{64.42}{1.32}
& \meanpm{4.30}{2.95} \\

Relearning-resilient
& - & - & - & - & - & - & - \\

Relearning-resilient + ALM 
& - & - & - & - & - & - & - \\

BLUR
& \meanpm{0.53}{0.11} & \meanpm{0.46}{0.09}
& \meanpm{55.30}{0.02}
& \meanpm{79.10}{0.02}
& \meanpm{75.40}{0.01}
& \meanpm{68.20}{0.02}
& \textcolor{red}{\meanpm{61.10}{0.04}} \\

PDU
& \textbf{\meanpm{0.53}{0.02}} & \meanpm{0.43}{0.06}
& \textbf{\meanpm{56.25}{0.47}}
& \underline{\meanpm{77.78}{1.05}}
& \meanpm{74.20}{2.59}
& \textbf{\meanpm{68.00}{1.13}}
& \meanpm{59.00}{2.03} \\

\midrule
\textbf{SAUL}
& \textbf{\meanpm{0.53}{0.01}} & \underline{\meanpm{0.03}{0.00}}
& \meanpm{52.34}{1.02}
& \textbf{\meanpm{78.46}{3.58}}
& \meanpm{65.00}{10.72}
& \meanpm{63.17}{3.58}
& \meanpm{1.10}{0.65} \\

w/o ALM
& \underline{\meanpm{0.51}{0.03}} & \textbf{\meanpm{0.02}{0.01}}
& \meanpm{51.44}{2.31}
& \meanpm{70.08}{10.32}
& \meanpm{46.80}{23.53}
& \meanpm{51.28}{15.36}
& \underline{\meanpm{0.30}{0.45}} \\

w/o SAM
& \textbf{\meanpm{0.53}{0.01}} & \underline{\meanpm{0.03}{0.01}}
& \meanpm{52.38}{1.91}
& \meanpm{75.02}{3.93}
& \meanpm{61.00}{13.62}
& \meanpm{60.66}{4.85}
& \underline{\meanpm{0.30}{0.27}} \\

w/o ALM + SAM
& \meanpm{0.50}{0.02} & \textbf{\meanpm{0.02}{0.01}}
& \meanpm{52.32}{1.67}
& \meanpm{68.74}{10.13}
& \meanpm{40.00}{16.58}
& \meanpm{49.36}{11.62}
& \underline{\meanpm{0.30}{0.27}} \\

\bottomrule
\end{tabular}
\label{tab:forget05_qpara}
\end{table*}

\clearpage

\begin{table*}[!t]
\centering
\small
\setlength{\tabcolsep}{3.0pt}
\renewcommand{\arraystretch}{0.95}
\caption{Unlearning performance with forget ratio = 10\% (same notation as Table~\ref{tab:forget1}).}
\label{tab:forget10}
\begin{tabular}{@{}l || cc|ccc|cc@{}}
\toprule
Method
& \multicolumn{2}{c|}{\textbf{Automatic}}
& \multicolumn{5}{c}{\textbf{GPT-based}} \\
\cmidrule(r){2-3}\cmidrule(l){4-8}
& MU$\uparrow$ & $F_{\text{ROUGE}}\downarrow$
& Retain$\uparrow$
& World Facts$\uparrow$
& Real Authors$\uparrow$
& HM$\uparrow$
& Forget$\downarrow$ \\
\midrule

Original Model
& 0.60 & 0.82 & 81.3 & 80.3 & 81.0 & 80.9 & 84.5 \\

\midrule

Single AdamW
& \meanpm{0.50}{0.01} & \textbf{\meanpm{0.00}{0.00}}
& \meanpm{54.64}{0.40}
& \meanpm{66.67}{0.60}
& \meanpm{39.00}{1.58}
& \meanpm{50.88}{0.98}
& \meanpm{0.50}{0.71} \\

Dual AdamW
& \meanpm{0.54}{0.00} & \underline{\meanpm{0.01}{0.01}}
& \meanpm{65.20}{0.84}
& \meanpm{76.07}{0.60}
& \meanpm{62.60}{1.67}
& \meanpm{67.47}{0.79}
& \meanpm{0.60}{0.89} \\

Dual AdamW + ALM
& \meanpm{0.55}{0.00} & \underline{\meanpm{0.01}{0.00}}
& \meanpm{65.40}{0.47}
& \meanpm{76.92}{0.60}
& \meanpm{65.40}{1.14}
& \meanpm{68.83}{0.65}
& \meanpm{0.40}{0.42} \\

Sharp Min--Max
& \meanpm{0.58}{0.00} & \meanpm{0.02}{0.00}
& \underline{\meanpm{85.44}{0.91}}
& \meanpm{75.38}{0.72}
& \underline{\meanpm{78.80}{0.84}}
& \underline{\meanpm{79.66}{0.46}}
& \underline{\meanpm{0.35}{0.34}} \\

Sharp Min--Max + ALM
& \underline{\meanpm{0.59}{0.00}} & \underline{\meanpm{0.01}{0.01}}
& \textbf{\meanpm{85.58}{0.43}}
& \underline{\meanpm{81.37}{0.72}}
& \textbf{\meanpm{79.40}{1.14}}
& \textbf{\meanpm{82.03}{0.45}}
& \textbf{\meanpm{0.30}{0.54}} \\

Relearning-resilient
& \meanpm{0.15}{0.00} & \underline{\meanpm{0.01}{0.01}}
& - & - & - & - & - \\

Relearning-resilient + ALM 
& \meanpm{0.17}{0.00} & \underline{\meanpm{0.01}{0.01}}
& - & - & - & - & - \\

BLUR
& \meanpm{0.58}{0.19} & \meanpm{0.33}{0.24}
& \meanpm{82.00}{0.15}
& \meanpm{82.00}{0.20}
& \meanpm{75.00}{0.42}
& \meanpm{79.50}{0.22}
& \textcolor{red}{\meanpm{28.75}{0.02}} \\

PDU
& \textbf{\meanpm{0.60}{0.02}} & \meanpm{0.03}{0.01}
& \meanpm{76.25}{0.66}
& \meanpm{80.34}{1.48}
& \meanpm{75.00}{3.46}
& \meanpm{77.09}{0.90}
& \meanpm{0.92}{0.38} \\

\midrule
\textbf{SAUL}
& \textbf{\meanpm{0.60}{0.00}} & \underline{\meanpm{0.01}{0.01}}
& \meanpm{71.08}{0.43}
& \textbf{\meanpm{82.22}{0.72}}
& \meanpm{78.20}{0.84}
& \meanpm{76.89}{0.55}
& \textbf{\meanpm{0.30}{0.45}} \\

w/o ALM
& \meanpm{0.57}{0.00} & \underline{\meanpm{0.01}{0.00}}
& \meanpm{64.10}{0.29}
& \meanpm{74.02}{1.67}
& \meanpm{68.80}{1.92}
& \meanpm{68.71}{0.50}
& \textbf{\meanpm{0.30}{0.27}} \\

w/o SAM
& \meanpm{0.55}{0.01} & \meanpm{0.02}{0.00}
& \meanpm{57.72}{0.81}
& \meanpm{72.65}{0.60}
& \meanpm{68.40}{2.70}
& \meanpm{65.61}{1.08}
& \meanpm{0.60}{0.55} \\

w/o ALM + SAM
& \meanpm{0.51}{0.00} & \textbf{\meanpm{0.00}{0.00}}
& \meanpm{52.24}{1.92}
& \meanpm{56.07}{0.76}
& \meanpm{49.20}{1.92}
& \meanpm{52.33}{1.12}
& \meanpm{0.80}{0.57} \\

\bottomrule
\end{tabular}
\end{table*}


\begin{table*}[!t]
\centering
\small
\setlength{\tabcolsep}{3.0pt}
\renewcommand{\arraystretch}{0.95}
\caption{Unlearning performance with forget ratio = 10\% under paraphrased question (same notation as Table~\ref{tab:forget1_qpara}).}
\label{tab:forget10_qpara}
\begin{tabular}{@{}l || cc|ccc|cc@{}}
\toprule
Method
& \multicolumn{2}{c|}{\textbf{Automatic}}
& \multicolumn{5}{c}{\textbf{GPT-based}} \\
\cmidrule(r){2-3}\cmidrule(l){4-8}
& MU$\uparrow$ & $F_{\text{ROUGE}}\downarrow$
& Retain$\uparrow$
& World Facts$\uparrow$
& Real Authors$\uparrow$
& HM$\uparrow$
& Forget$\downarrow$ \\
\midrule

Original Model
& 0.55 & 0.48 & 56.8 & 79.5 & 77.0 & 69.5 & 67.0 \\

\midrule

Single AdamW
& \meanpm{0.43}{0.02} & \underline{\meanpm{0.02}{0.01}}
& \meanpm{42.54}{2.04}
& \meanpm{67.86}{4.96}
& \meanpm{26.20}{4.44}
& \meanpm{38.95}{3.23}
& \meanpm{1.56}{1.48} \\

Dual AdamW
& \meanpm{0.44}{0.16} & \textbf{\meanpm{0.01}{0.01}}
& \meanpm{49.00}{4.04}
& \meanpm{62.74}{28.51}
& \meanpm{59.40}{11.46}
& \meanpm{52.99}{3.80}
& \underline{\meanpm{0.46}{0.62}} \\

Dual AdamW + ALM
& \textbf{\meanpm{0.53}{0.01}} & \meanpm{0.03}{0.01}
& \underline{\meanpm{52.32}{2.33}}
& \meanpm{78.62}{4.22}
& \meanpm{73.20}{6.50}
& \meanpm{60.86}{2.17}
& \meanpm{1.80}{0.92} \\

Sharp Min--Max
& \meanpm{0.45}{0.06} & \meanpm{0.05}{0.01}
& \meanpm{46.82}{1.10}
& \meanpm{69.58}{4.04}
& \meanpm{34.00}{12.73}
& \meanpm{38.05}{10.21}
& \meanpm{3.46}{2.11} \\

Sharp Min--Max + ALM
& \meanpm{0.49}{0.04} & \meanpm{0.13}{0.04}
& \meanpm{47.76}{0.34}
& \meanpm{71.40}{1.67}
& \meanpm{41.60}{1.67}
& \meanpm{50.84}{0.91}
& \meanpm{1.00}{0.47} \\

Relearning-resilient
& - & - & - & - & - & - & - \\

Relearning-resilient + ALM 
& - & - & - & - & - & - & - \\

BLUR
& \meanpm{0.54}{0.06} & \meanpm{0.28}{0.10}
& \meanpm{60.00}{0.16}
& \meanpm{79.00}{0.53}
& \meanpm{72.00}{0.10}
& \meanpm{69.40}{0.43}
& \textcolor{red}{\meanpm{27.00}{0.14}} \\

PDU
& \textbf{\meanpm{0.53}{0.00}} & \meanpm{0.03}{0.01}
& \textbf{\meanpm{56.75}{1.39}}
& \meanpm{74.64}{0.99}
& \meanpm{70.00}{1.73}
& \textbf{\meanpm{66.20}{0.71}}
& \meanpm{1.25}{0.43} \\

\midrule

\textbf{SAUL}
& \textbf{\meanpm{0.53}{0.01}} & \meanpm{0.03}{0.01}
& \meanpm{51.74}{5.19}
& \textbf{\meanpm{81.18}{2.00}}
& \underline{\meanpm{73.40}{7.27}}
& \underline{\meanpm{66.04}{3.78}}
& \meanpm{1.48}{0.93} \\

w/o ALM
& \meanpm{0.50}{0.03} & \textbf{\meanpm{0.01}{0.01}}
& \meanpm{49.20}{4.23}
& \meanpm{75.04}{8.86}
& \meanpm{57.20}{17.80}
& \meanpm{57.41}{9.62}
& \textbf{\meanpm{0.36}{0.57}} \\

w/o SAM
& \underline{\meanpm{0.52}{0.01}} & \meanpm{0.04}{0.04}
& \meanpm{48.32}{2.95}
& \underline{\meanpm{79.84}{2.76}}
& \textbf{\meanpm{75.20}{7.79}}
& \meanpm{64.31}{2.82}
& \meanpm{3.60}{4.87} \\

w/o ALM + SAM
& \meanpm{0.51}{0.02} & \textbf{\meanpm{0.01}{0.01}}
& \meanpm{50.74}{4.43}
& \meanpm{71.30}{5.82}
& \meanpm{54.60}{10.53}
& \meanpm{57.25}{5.82}
& \textbf{\meanpm{0.36}{0.80}} \\

\bottomrule
\end{tabular}
\end{table*}

\clearpage


\begin{table*}[!t]
\centering
\small
\setlength{\tabcolsep}{3.0pt}
\renewcommand{\arraystretch}{0.95}
\caption{Unlearning performance of LLaMA-3.2-3B with forget ratio = 1\% (same notation as Table~\ref{tab:forget1}).}
\label{tab:3b_forget01_qpara}
\begin{tabular}{@{}l || cc|ccc|cc@{}}
\toprule
Method
& \multicolumn{2}{c|}{\textbf{Automatic}}
& \multicolumn{5}{c}{\textbf{GPT-based}} \\
\cmidrule(r){2-3}\cmidrule(l){4-8}
& MU$\uparrow$ & $F_{\text{ROUGE}}\downarrow$
& Retain$\uparrow$
& World Facts$\uparrow$
& Real Authors$\uparrow$
& HM$\uparrow$
& Forget$\downarrow$ \\
\midrule

Original Model
& 0.67 & 0.99 & 94.0 & 88.0 & 88.0 & 89.91 & 100 \\

\midrule

Single AdamW
& \meanpm{0.63}{0.03} & \meanpm{0.02}{0.00}
& \meanpm{73.10}{1.68}
& \meanpm{62.80}{5.12}
& \meanpm{81.30}{2.18}
& \meanpm{71.59}{1.86}
& \meanpm{6.40}{0.34} \\

Dual AdamW
& \underline{\meanpm{0.65}{0.05}} & \underline{\meanpm{0.01}{0.01}}
& \meanpm{72.40}{2.94}
& \meanpm{67.40}{4.05}
& \meanpm{79.80}{1.66}
& \meanpm{72.85}{1.19}
& \meanpm{3.80}{0.15} \\

Dual AdamW + ALM
& \underline{\meanpm{0.65}{0.08}} & \underline{\meanpm{0.01}{0.04}}
& \meanpm{73.50}{2.42}
& \meanpm{68.10}{2.91}
& \meanpm{78.40}{1.91}
& \meanpm{73.09}{0.50}
& \underline{\meanpm{2.70}{0.29}} \\

Sharp Min--Max
& \meanpm{0.59}{0.15} & \meanpm{0.03}{0.03}
& \meanpm{75.80}{1.96}
& \meanpm{72.10}{1.91}
& \meanpm{81.20}{2.08}
& \meanpm{76.19}{0.09}
& \meanpm{5.40}{0.18} \\

Sharp Min--Max + ALM
& \meanpm{0.60}{0.06} & \meanpm{0.03}{0.04}
& \meanpm{74.20}{4.80}
& \meanpm{73.50}{2.94}
& \meanpm{81.40}{3.84}
& \meanpm{76.20}{0.93}
& \meanpm{3.40}{0.34} \\

Relearning-resilient
& -- & -- & -- & -- & -- & -- & -- \\

Relearning-resilient + ALM
& -- & -- & -- & -- & -- & -- & -- \\

\midrule

\textbf{SAUL}
& \textbf{\meanpm{0.66}{0.43}} & \meanpm{0.02}{0.02}
& \textbf{\meanpm{78.60}{3.36}}
& \textbf{\meanpm{85.40}{1.42}}
& \underline{\meanpm{82.00}{2.34}}
& \textbf{\meanpm{82.24}{0.97}}
& \textbf{\meanpm{2.20}{0.27}} \\

w/o ALM
& \meanpm{0.63}{0.06} & \textbf{\meanpm{0.00}{0.05}}
& \underline{\meanpm{76.70}{2.64}}
& \underline{\meanpm{81.60}{0.99}}
& \meanpm{79.50}{3.43}
& \meanpm{79.22}{1.25}
& \meanpm{2.80}{0.20} \\

w/o SAM
& \textbf{\meanpm{0.66}{0.05}} & \underline{\meanpm{0.01}{0.06}}
& \meanpm{76.40}{3.12}
& \meanpm{80.60}{2.46}
& \textbf{\meanpm{82.70}{4.54}}
& \underline{\meanpm{79.81}{1.06}}
& \meanpm{2.90}{0.22} \\

w/o ALM + SAM
& \textbf{\meanpm{0.66}{0.47}} & \underline{\meanpm{0.01}{0.02}}
& \meanpm{72.50}{2.95}
& \meanpm{65.40}{3.45}
& \meanpm{77.40}{1.92}
& \meanpm{71.42}{0.78}
& \meanpm{5.20}{0.23} \\

\bottomrule
\end{tabular}
\end{table*}


\begin{table*}[!t]
\centering
\small
\setlength{\tabcolsep}{3.0pt}
\renewcommand{\arraystretch}{0.95}
\caption{Unlearning performance of LLaMA-3.2-3B with forget ratio = 5\% (same notation as Table~\ref{tab:forget1}).}
\label{tab:3b_forget05_qpara}
\begin{tabular}{@{}l || cc|ccc|cc@{}}
\toprule
Method
& \multicolumn{2}{c|}{\textbf{Automatic}}
& \multicolumn{5}{c}{\textbf{GPT-based}} \\
\cmidrule(r){2-3}\cmidrule(l){4-8}
& MU$\uparrow$ & $F_{\text{ROUGE}}\downarrow$
& Retain$\uparrow$
& World Facts$\uparrow$
& Real Authors$\uparrow$
& HM$\uparrow$
& Forget$\downarrow$ \\
\midrule

Original Model
& 0.68 & 0.94 & 94.0 & 88.0 & 88.0 & 89.91 & 95.0 \\

\midrule

Single AdamW
& \meanpm{0.61}{0.04} & \underline{\meanpm{0.02}{0.02}}
& \meanpm{69.70}{5.42}
& \meanpm{61.40}{5.31}
& \meanpm{78.50}{1.43}
& \meanpm{69.17}{2.80}
& \meanpm{2.25}{0.16} \\

Dual AdamW
& \meanpm{0.63}{0.16} & \underline{\meanpm{0.02}{0.03}}
& \meanpm{72.40}{4.31}
& \meanpm{69.40}{1.92}
& \meanpm{83.10}{2.41}
& \meanpm{74.52}{2.57}
& \textbf{\meanpm{0.43}{0.05}} \\

Dual AdamW + ALM
& \underline{\meanpm{0.64}{0.04}} & \textbf{\meanpm{0.00}{0.04}}
& \meanpm{73.50}{8.13}
& \meanpm{72.10}{0.84}
& \meanpm{82.70}{3.81}
& \meanpm{75.82}{1.90}
& \meanpm{0.75}{0.75} \\

Sharp Min--Max
& \meanpm{0.62}{0.07} & \underline{\meanpm{0.02}{0.08}}
& \meanpm{71.70}{2.46}
& \meanpm{68.40}{1.37}
& \meanpm{84.30}{1.61}
& \meanpm{74.20}{1.71}
& \meanpm{1.25}{0.04} \\

Sharp Min--Max + ALM
& \underline{\meanpm{0.64}{0.10}} & \meanpm{0.03}{0.07}
& \meanpm{74.60}{0.46}
& \meanpm{71.60}{2.46}
& \textbf{\meanpm{85.40}{0.93}}
& \meanpm{76.76}{0.82}
& \meanpm{0.54}{0.02} \\

Relearning-resilient
& -- & -- & -- & -- & -- & -- & -- \\

Relearning-resilient + ALM
& -- & -- & -- & -- & -- & -- & -- \\

\midrule

\textbf{SAUL}
& \textbf{\meanpm{0.65}{0.02}} & \meanpm{0.04}{0.04}
& \textbf{\meanpm{77.50}{1.41}}
& \underline{\meanpm{84.00}{3.40}}
& \underline{\meanpm{84.50}{2.41}}
& \textbf{\meanpm{81.87}{2.12}}
& \underline{\meanpm{0.46}{0.16}} \\

w/o ALM
& \meanpm{0.62}{0.15} & \underline{\meanpm{0.02}{0.01}}
& \underline{\meanpm{76.40}{1.44}}
& \textbf{\meanpm{84.30}{1.61}}
& \meanpm{84.30}{1.61}
& \underline{\meanpm{81.49}{1.55}}
& \meanpm{1.24}{0.01} \\

w/o SAM
& \underline{\meanpm{0.64}{0.09}} & \meanpm{0.09}{0.04}
& \meanpm{74.80}{2.41}
& \meanpm{83.70}{2.63}
& \meanpm{83.70}{0.95}
& \meanpm{80.51}{1.62}
& \meanpm{1.75}{0.09} \\

w/o ALM + SAM
& \meanpm{0.62}{0.04} & \meanpm{0.03}{0.03}
& \meanpm{68.40}{2.64}
& \meanpm{74.50}{1.06}
& \meanpm{79.50}{0.41}
& \meanpm{73.85}{0.80}
& \meanpm{0.75}{0.16} \\

\bottomrule
\end{tabular}
\end{table*}


\begin{table*}[!t]
\centering
\small
\setlength{\tabcolsep}{3.0pt}
\renewcommand{\arraystretch}{0.95}
\caption{Unlearning performance of LLaMA-3.2-3B with forget ratio = 10\% (same notation as Table~\ref{tab:forget1}).}
\label{tab:3b_forget10_qpara}
\begin{tabular}{@{}l || cc|ccc|cc@{}}
\toprule
Method
& \multicolumn{2}{c|}{\textbf{Automatic}}
& \multicolumn{5}{c}{\textbf{GPT-based}} \\
\cmidrule(r){2-3}\cmidrule(l){4-8}
& MU$\uparrow$ & $F_{\text{ROUGE}}\downarrow$
& Retain$\uparrow$
& World Facts$\uparrow$
& Real Authors$\uparrow$
& HM$\uparrow$
& Forget$\downarrow$ \\
\midrule

Original Model
& 0.67 & 0.98 & 93.25 & 88.0 & 88.0 & 89.68 & 95.75 \\

\midrule

Single AdamW
& \underline{\meanpm{0.63}{0.06}} & \textbf{\meanpm{0.00}{0.04}}
& \meanpm{73.10}{1.51}
& \meanpm{64.00}{1.09}
& \meanpm{85.40}{0.98}
& \meanpm{73.14}{1.15}
& \meanpm{1.75}{0.02} \\

Dual AdamW
& \underline{\meanpm{0.63}{0.07}} & \meanpm{0.03}{0.02}
& \meanpm{70.70}{0.54}
& \meanpm{68.00}{2.80}
& \meanpm{87.40}{0.54}
& \meanpm{74.46}{0.74}
& \underline{\meanpm{1.40}{0.00}} \\

Dual AdamW + ALM
& \underline{\meanpm{0.63}{0.10}} & \textbf{\meanpm{0.00}{0.02}}
& \underline{\meanpm{74.50}{0.85}}
& \meanpm{70.00}{0.50}
& \textbf{\meanpm{90.59}{0.81}}
& \meanpm{77.42}{0.68}
& \meanpm{6.00}{0.02} \\

Sharp Min--Max
& \meanpm{0.61}{0.15} & \underline{\meanpm{0.01}{0.05}}
& \meanpm{68.40}{0.91}
& \meanpm{67.00}{0.91}
& \meanpm{88.10}{1.64}
& \meanpm{73.36}{1.07}
& \meanpm{4.00}{0.04} \\

Sharp Min--Max + ALM
& \underline{\meanpm{0.63}{0.07}} & \meanpm{0.03}{0.00}
& \meanpm{73.10}{1.72}
& \meanpm{74.00}{1.34}
& \meanpm{87.90}{2.50}
& \underline{\meanpm{77.78}{1.74}}
& \textbf{\meanpm{1.20}{0.01}} \\

Relearning-resilient
& -- & -- & -- & -- & -- & -- & -- \\

Relearning-resilient + ALM
& -- & -- & -- & -- & -- & -- & -- \\

\midrule

\textbf{SAUL}
& \textbf{\meanpm{0.65}{0.03}} & \meanpm{0.02}{0.02}
& \textbf{\meanpm{78.50}{1.12}}
& \textbf{\meanpm{86.00}{2.41}}
& \meanpm{87.90}{1.24}
& \textbf{\meanpm{83.93}{1.42}}
& \meanpm{2.34}{0.05} \\

w/o ALM
& \meanpm{0.58}{0.04} & \meanpm{0.02}{0.01}
& \meanpm{61.75}{1.41}
& \underline{\meanpm{81.00}{2.83}}
& \underline{\meanpm{89.70}{0.19}}
& \meanpm{75.59}{0.47}
& \meanpm{5.00}{0.04} \\

w/o SAM
& \meanpm{0.57}{0.04} & \underline{\meanpm{0.01}{0.12}}
& \meanpm{62.30}{1.23}
& \meanpm{75.00}{1.93}
& \meanpm{86.30}{0.54}
& \meanpm{73.22}{0.94}
& \meanpm{4.86}{0.04} \\

w/o ALM + SAM
& \meanpm{0.57}{0.05} & \underline{\meanpm{0.01}{0.15}}
& \meanpm{53.00}{0.98}
& \meanpm{57.00}{1.16}
& \meanpm{82.05}{1.84}
& \meanpm{61.73}{1.24}
& \meanpm{5.84}{0.08} \\

\bottomrule
\end{tabular}
\end{table*}

\clearpage
\subsection{Results on MUSE News}
\label{app:muse_additional}

\begin{table}[H]
\centering
\small
\setlength{\tabcolsep}{4pt}
\renewcommand{\arraystretch}{1.0}
\caption{Additional MUSE News results using \texttt{Llama-2-7B}.
Original denotes the model trained on the full dataset, and Retrain denotes
the model retrained after excluding the forget set $\mathcal{D}_f$.
Results for the unlearning methods are reported as mean $\pm$ standard
deviation over three random seeds.}
\label{tab:muse_news_additional}
\begin{adjustbox}{max width=\columnwidth}
\begin{tabular}{@{}lcccc@{}}
\toprule
Method
& \makecell[c]{VerbMem\\$\downarrow$}
& \makecell[c]{KnowMem\\$\mathcal{D}_f \downarrow$}
& \makecell[c]{PrivLeak\\$\rightarrow 0$}
& \makecell[c]{KnowMem\\$\mathcal{D}_r \uparrow$} \\
\midrule
Original
& 58.4 & 63.9 & -99.8 & 55.2 \\
Retrain
& 20.8 & 33.1 & 0.0 & 55.0 \\
\midrule
Sharp Min--Max
& 0.03 $\pm$ 0.05
& 30.84 $\pm$ 1.87
& -96.50 $\pm$ 1.65
& 25.25 $\pm$ 0.47 \\
Relearning-resilient
& 0.00 $\pm$ 0.00
& 32.01 $\pm$ 1.28
& 109.42 $\pm$ 0.19
& 29.49 $\pm$ 0.30 \\
BLUR
& 0.73 $\pm$ 0.81
& 31.57 $\pm$ 1.53
& 105.79 $\pm$ 3.22
& 31.02 $\pm$ 3.41 \\
SAUL
& 9.08 $\pm$ 1.33
& 32.63 $\pm$ 6.22
& 61.95 $\pm$ 9.42
& 30.19 $\pm$ 2.73 \\
\bottomrule
\end{tabular}
\end{adjustbox}
\end{table}

\clearpage
\section{Computational Overhead of SAUL}
\label{app:computational_overhead}

SAM introduces additional computational overhead because it requires extra forward and backward passes at each iteration. To make this cost--benefit trade-off explicit, we report the runtime and peak memory usage of SAUL relative to representative baselines on ToFU with a forget ratio of $10\%$ using \texttt{LLaMA-3.2-3B}. The results are shown in Tables~\ref{tab:runtime_overhead} and~\ref{tab:memory_overhead}.

\begin{table}[ht]
\centering
\small
\setlength{\tabcolsep}{6pt}
\renewcommand{\arraystretch}{1.05}
\caption{Runtime comparison on ToFU with a forget ratio of $10\%$ using \texttt{LLaMA-3.2-3B}.}
\begin{tabular}{lcccc}
\toprule
Method & SAUL & SAUL w/o SAM & Dual-AdamW & Sharp Min--Max \\
\midrule
Time (s) & 267 & 221 & 164 & 232 \\
\bottomrule
\end{tabular}
\label{tab:runtime_overhead}
\end{table}

\begin{table}[ht]
\centering
\small
\setlength{\tabcolsep}{8pt}
\renewcommand{\arraystretch}{1.05}
\caption{Peak memory usage on ToFU with a forget ratio of $10\%$ using \texttt{LLaMA-3.2-3B}.}
\begin{tabular}{lcc}
\toprule
Method & SAUL & Dual-AdamW \\
\midrule
Peak Memory (MB) & 86459.18 & 81834.65 \\
\bottomrule
\end{tabular}
\label{tab:memory_overhead}
\end{table}

SAUL requires $267$ seconds, compared to $221$ seconds for SAUL w/o SAM, $164$ seconds for Dual-AdamW, and $232$ seconds for Sharp Min--Max. Peak memory usage increases from $79.92$ GB for Dual-AdamW to $84.43$ GB for SAUL. This additional cost is accompanied by improved robustness under query rephrasing, where SAUL improves the GPT-paraphrased ToFU harmonic mean from $64.31$ to $66.04$ compared to its w/o SAM variant using \texttt{LLaMA-3.2-1B}.

\clearpage
\section{Threshold Calibration via Margin-Based Lower Bounds}
\label{sec:threshold-calibration}

The threshold $\alpha$ in SAUL denotes a prescribed satisfaction
level for the chosen forget-side constraint. In the main implementation, the constraint is defined using the cross-entropy-based forget loss $\mathcal{L}_f$, so the numerical value of $\alpha$ is measured on the scale of that loss.

In this appendix, we consider a margin-based instantiation in which the constraint quantity is a prediction-level certificate $\Phi_f$.
For notational consistency, we retain the symbol $\alpha$ for the corresponding threshold; however, its numerical value is specific to the chosen constraint quantity and need not coincide with the loss-scale threshold used in the main implementation.
Under this margin-based instantiation, $\alpha\in[0,1]$ admits a direct interpretation as a target forget rate.

The key idea is asymmetric: on the retain side, cross-entropy is used as an upper surrogate for token-level $0/1$ error, whereas on the forget side, a margin-based lower surrogate is maximized to certify sequence-level exact-match failure.

\subsection{Prediction margin and token-level 0/1 error}
\label{subsec:margin}

For a categorical predictor with output distribution $p_\theta(y\mid x)$ over a finite label space, define the prediction margin
\begin{equation}
m(\theta;x,y)
:=
p_\theta(y\mid x)
-
\max_{y'\neq y} p_\theta(y'\mid x).
\label{eq:prediction-margin}
\end{equation}
The target label is the unique top-1 prediction exactly when $m(\theta;x,y)>0$.
Thus, the token-level $0/1$ error can be written as
\begin{equation}
\ell_{0\text{-}1}(\theta;x,y)
=
\mathbf{1}\!\left[m(\theta;x,y)\le 0\right].
\label{eq:zero-one-margin}
\end{equation}
This margin-based formulation directly applies to autoregressive generation, where each next-token prediction defines a categorical prediction problem over the vocabulary.

For an answer sequence $a=(a_1,\ldots,a_T)$ conditioned on a prompt $q$, we define the token-level margin
\begin{equation}
m_t(\theta;q,a)
:=
p_\theta(a_t\mid q,a_{<t})
-
\max_{v\neq a_t}p_\theta(v\mid q,a_{<t}).
\label{eq:token-margin}
\end{equation}
The token is predicted incorrectly whenever $m_t(\theta;q,a)\le 0$.

\subsection{Retain side: cross-entropy as an upper surrogate}
\label{subsec:retain-upper}

The cross-entropy loss
\begin{equation}
\ell_{\mathrm{CE}}(\theta;x,y)
:=
-\log p_\theta(y\mid x)
\end{equation}
upper-bounds the token-level $0/1$ error up to the constant $1/\log 2$:
\begin{equation}
\ell_{0\text{-}1}(\theta;x,y)
\le
\frac{1}{\log 2}\ell_{\mathrm{CE}}(\theta;x,y).
\label{eq:ce-upper-bound}
\end{equation}
Indeed, if the top-1 prediction is incorrect, then some non-target class has probability at least $p_\theta(y\mid x)$, which implies $p_\theta(y\mid x)\le 1/2$ and hence $\ell_{\mathrm{CE}}(\theta;x,y)\ge \log 2$.
If the prediction is correct, the left-hand side is zero.

For a retain set $\mathcal{D}_r$, define the length-normalized retain loss
\begin{equation}
\mathcal{L}_r(\theta)
:=
\frac{1}{|\mathcal{D}_r|}
\sum_{(q,a)\in\mathcal{D}_r}
\frac{1}{T_{q,a}}
\sum_{t=1}^{T_{q,a}}
-\log p_\theta(a_t\mid q,a_{<t}),
\label{eq:retain-ce}
\end{equation}
where $T_{q,a}$ denotes the length of the target answer $a$.
Averaging~\eqref{eq:ce-upper-bound} token-by-token gives
\begin{equation}
\mathrm{TokErr}_r(\theta)
\le
\frac{1}{\log 2}\mathcal{L}_r(\theta),
\label{eq:retain-token-bound}
\end{equation}
where
\begin{equation}
\mathrm{TokErr}_r(\theta)
:=
\frac{1}{|\mathcal{D}_r|}
\sum_{(q,a)\in\mathcal{D}_r}
\frac{1}{T_{q,a}}
\sum_{t=1}^{T_{q,a}}
\mathbf{1}\!\left[m_t(\theta;q,a)\le 0\right].
\end{equation}
Thus, minimizing $\mathcal{L}_r$ controls a computable upper bound on retain-side token error.

\subsection{Forget side: a lower surrogate for exact-match failure}
\label{subsec:forget-lower}

On the forget side, the goal is reversed: we want the model to fail to reproduce the target answer.
For generation tasks, this is naturally measured by sequence-level exact-match error,
\begin{equation}
\mathrm{EM}(\theta;q,a)
:=
\mathbf{1}\!\left[\exists\,t:\;m_t(\theta;q,a)\le 0\right],
\label{eq:sequence-em}
\end{equation}
which equals one as soon as at least one target token is not the unique top-1 prediction.

To obtain a differentiable certificate for this error, we use a smooth margin-based lower surrogate.
Let $\tau>0$ be a fixed sharpness parameter and define
\begin{equation}
\phi_\tau(m)
:=
1-\log\!\left(1+\exp\!\left(\frac{m}{\tau}+1\right)\right).
\label{eq:margin-lower-surrogate}
\end{equation}

\begin{figure}[t]
\centering
\includegraphics[width=1.0\linewidth]{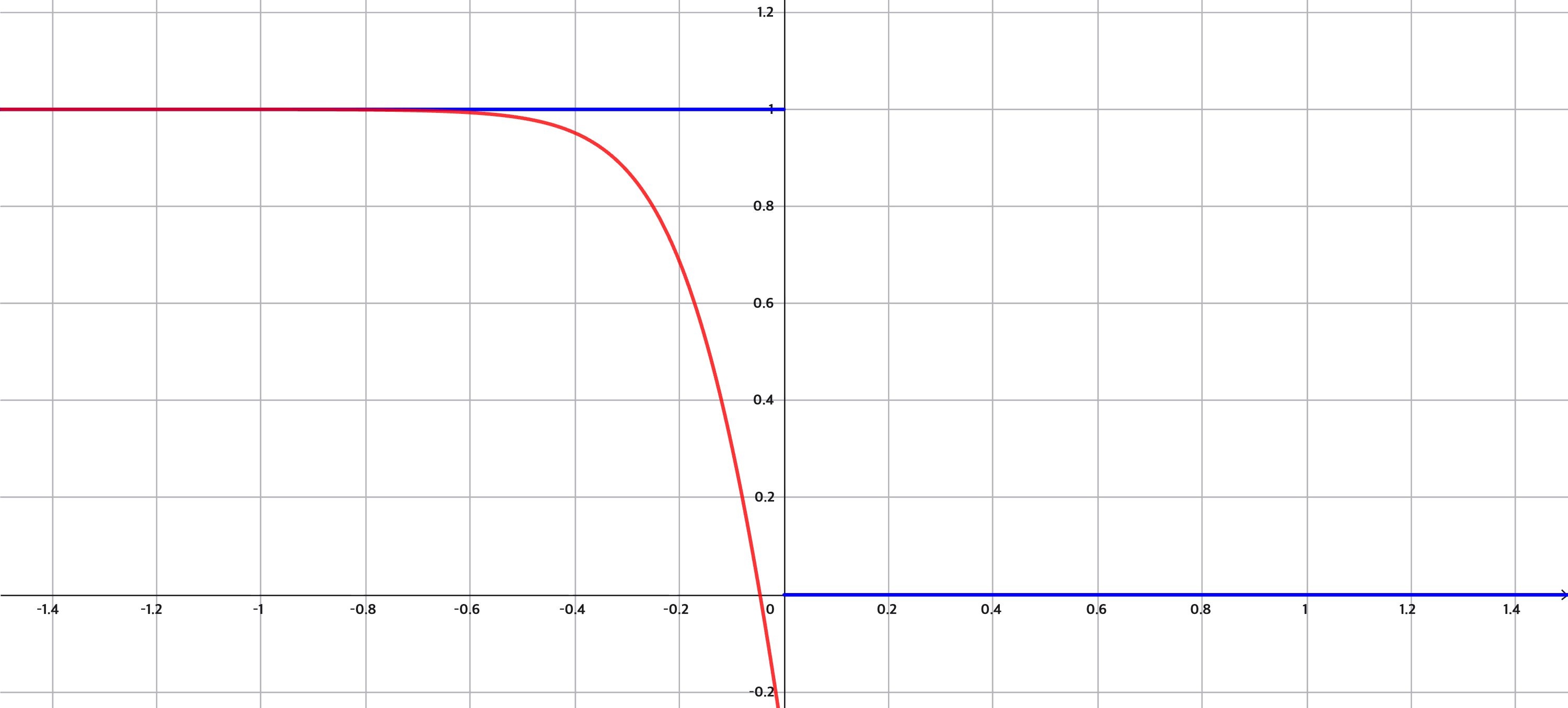}
\caption{
Margin-based lower surrogate for token-level prediction failure with $\tau=0.1$.
The blue step function denotes the token-level $0/1$ error $\mathbf{1}[m\le 0]$, and the red curve denotes the smooth lower surrogate $\phi_\tau(m)$.
The surrogate remains below the $0/1$ error for all margins while approaching one for sufficiently negative margins.
}
\label{fig:margin-lower-bound}
\end{figure}

As illustrated in Figure~\ref{fig:margin-lower-bound}, the red surrogate curve with $\tau=0.1$ provides a smooth lower bound on the blue token-level $0/1$ error curve.
The parameter $\tau$ controls the sharpness of the surrogate: smaller $\tau$ makes $\phi_\tau$ sharper around the decision boundary $m=0$, while larger $\tau$ yields a smoother curve.
Although the pointwise lower-bound property holds for any $\tau>0$, $\tau$ affects the numerical scale and tightness of the certificate.
In particular, within the feasible probability-margin range $m\in[-1,1]$, $\tau$ should be small enough for negative margins to produce surrogate values close to one.
For example, $\phi_1(-1)=1-\log 2\approx 0.307$, whereas $\phi_{0.1}(-1)\approx 1$.

The surrogate is monotone decreasing in $m$, rewards negative margins, and penalizes positive margins.
Moreover, it is a pointwise lower bound on the token-level $0/1$ error:
\begin{equation}
\phi_\tau(m)
\le
\mathbf{1}[m\le 0]
\qquad
\text{for all }m.
\label{eq:pointwise-lower-bound}
\end{equation}
If $m>0$, then $m/\tau+1>1$, so $\phi_\tau(m)<1-\log(1+e)<0=\mathbf{1}[m\le 0]$.
If $m\le 0$, then $\phi_\tau(m)\le 1=\mathbf{1}[m\le 0]$.
Therefore~\eqref{eq:pointwise-lower-bound} holds for all margins.

We aggregate this lower surrogate at the sequence level by taking the strongest token-level evidence of failure:
\begin{equation}
\Phi_f(\theta)
:=
\frac{1}{|\mathcal{D}_f|}
\sum_{(q,a)\in\mathcal{D}_f}
\max_{t\in\{1,\ldots,T_{q,a}\}}
\phi_\tau\!\left(m_t(\theta;q,a)\right).
\label{eq:forget-certificate}
\end{equation}
Since $\phi_\tau(m_t)\le \mathbf{1}[m_t\le 0]$ for every token, the maximum also preserves the lower-bound relation:
\begin{equation}
\max_t \phi_\tau(m_t(\theta;q,a))
\le
\mathbf{1}\!\left[\exists\,t:\;m_t(\theta;q,a)\le 0\right].
\end{equation}
Averaging over the forget set gives
\begin{equation}
\Phi_f(\theta)
\le
\mathrm{EM}_f(\theta),
\label{eq:forget-em-lower-bound}
\end{equation}
where
\begin{equation}
\mathrm{EM}_f(\theta)
:=
\frac{1}{|\mathcal{D}_f|}
\sum_{(q,a)\in\mathcal{D}_f}
\mathrm{EM}(\theta;q,a).
\end{equation}
Consequently, within this margin-based instantiation, enforcing
\begin{equation}
\Phi_f(\theta)\ge \alpha
\label{eq:forget-alpha-certificate}
\end{equation}
certifies
\begin{equation}
\mathrm{EM}_f(\theta)\ge \alpha.
\label{eq:certified-forgetting}
\end{equation}
For example, $\alpha=0.95$ certifies that at least $95\%$ of forget examples fail exact match, provided that the chosen $\tau$ makes this certificate numerically attainable.

\subsection{Certified SAUL formulation and interpretation of
\texorpdfstring{$\alpha$}{alpha}}
\label{subsec:saul-alpha}

The prediction-level certificate above motivates the following
certified instantiation of the SAUL framework:
\begin{equation}
\min_{\theta}\;\mathcal{L}_r(\theta)
\qquad
\text{s.t.}
\qquad
\Phi_f(\theta)\ge \alpha.
\label{eq:saul-certified-form}
\end{equation}
The retain objective minimizes an upper surrogate of token-level retain error, while the forget constraint maximizes a lower surrogate of sequence-level exact-match failure.
This forms an upper/lower surrogate pair around the two asymmetric goals of unlearning: preserve retain-side correctness and induce forget-side prediction failure.

Under this certified formulation, $\alpha$ is not a relative
trade-off weight between forgetting and retention.
It is a prediction-level target: if the constraint is satisfied,
then~\eqref{eq:certified-forgetting} guarantees that the exact-match
error on the forget set is at least $\alpha$.
Thus, choosing $\alpha=0.95$ directly encodes the requirement that
at least $95\%$ of forget examples should fail exact match.
This differs from scalarized objectives, where the desired level of
forgetting is specified implicitly through optimization
coefficients.

\paragraph{Relation to cross-entropy-based forget losses.}
When the implemented forget-side optimizer uses a
cross-entropy-based objective, the margin certificate
$\Phi_f(\theta)$ can be evaluated post hoc rather than optimized
directly.
Increasing the forget-side cross-entropy decreases the target-token
probability and therefore tends to decrease the margin
$m_t(\theta;q,a)$, which increases $\phi_\tau(m_t)$.
Thus, the cross-entropy forget objective and the margin-based lower
surrogate are directionally aligned, although they are not identical
functions.

This directional alignment does not imply that the threshold values
defined on the two constraint scales are numerically identical.
Instead, a model trained using a cross-entropy-based threshold can be
evaluated against a desired prediction-level target $\alpha$ by
measuring whether the resulting model satisfies
$\Phi_f(\theta)\ge\alpha$.

\subsection{Augmented-Lagrangian gating}
\label{subsec:hinge-surrogate}

SAUL enforces the forget constraint through an augmented-Lagrangian update.
For the certified formulation in~\eqref{eq:saul-certified-form}, the constraint quantity is $\Phi_f$.
In our implementation, we use the cross-entropy-based forget loss $\mathcal{L}_f$ as a tractable proxy for this constraint quantity, while using $\Phi_f$ as the calibration certificate.
Since increasing $\mathcal{L}_f$ decreases the target-token probability and hence decreases the margin, $\mathcal{L}_f$ and $\Phi_f$ are directionally aligned.
Thus, the same gating logic applies with $\mathcal{L}_f$ in place of $\Phi_f$ during training, and $\alpha$ is validated by measuring whether the resulting model satisfies $\Phi_f(\theta)\ge\alpha$.

Writing the certified constraint violation as
\begin{equation}
g(\theta)
:=
\alpha-\Phi_f(\theta),
\end{equation}
the multiplier update takes the projected form
\begin{equation}
\lambda^+
\leftarrow
\left[\lambda+\mu g(\theta)\right]_+.
\label{eq:alm-projection}
\end{equation}
This projection implements a hinge-style gating mechanism.
When $\Phi_f(\theta)\ge\alpha$, the constraint is satisfied and
$g(\theta)\le 0$, so the projected update decreases the multiplier.
If the criterion remains satisfied, the multiplier can eventually
reach zero, at which point the forget-side pressure is deactivated.
When $\Phi_f(\theta)<\alpha$, the constraint is violated and
$g(\theta)>0$, so the multiplier increases and strengthens the
forget-side pressure in proportion to the violation.

The augmented-Lagrangian term combines this adaptive gating with a quadratic penalty on the violated constraint:
\begin{equation}
\lambda^+(\alpha-\Phi_f(\theta))
+
\frac{\mu}{2}
\left(\alpha-\Phi_f(\theta)\right)^2.
\label{eq:alm-term}
\end{equation}
The projection $[\cdot]_+$ provides the hinge-like deactivation behavior, while the quadratic term stabilizes multiplier dynamics.
Therefore, under the certified instantiation, the controller
increases forget-side pressure when the certified lower bound on
exact-match failure falls below $\alpha$ and reduces it as the
criterion remains satisfied.
In the cross-entropy implementation, the same gating logic is
applied using the corresponding loss-scale constraint.

\clearpage
\section{Broader Impact and Potential Risks}
\label{app:impact_statement}

LLM unlearning aims to reduce privacy, copyright, and misuse risks by removing the influence of designated target data while preserving general model utility.
SAUL contributes to this goal by treating forgetting as an explicit constraint, so that optimization can stop applying forget-side pressure once a prescribed forgetting criterion is satisfied.

At the same time, unlearning methods can create risks if their effectiveness is overinterpreted.
A model may satisfy benchmark-level forgetting criteria while still revealing target knowledge under stronger prompting, relearning, or recovery procedures.
Therefore, SAUL should not be interpreted as providing a complete removal guarantee against all post-unlearning attacks.
Its empirical results should be understood within the evaluated benchmarks, metrics, and threat models.

Another risk is over-forgetting.
If the forgetting threshold is set too aggressively, an unlearning method may degrade benign capabilities or remove related non-target knowledge.
This is especially important for LLMs, where target knowledge can be entangled with surrounding factual, linguistic, or domain knowledge.
The constrained formulation in SAUL is designed to reduce unnecessary forgetting pressure, but appropriate threshold selection and post-unlearning evaluation remain necessary.

We do not release new sensitive datasets or model checkpoints as part of this work.
Our experiments use existing research benchmarks and report aggregate evaluation results.
Future deployments of unlearning methods should be accompanied by task-specific risk assessment, stronger adaptive recovery evaluations, and validation that the resulting model does not introduce unacceptable degradation on benign use cases.

\end{document}